\documentclass[11pt]{article}

\usepackage[utf8]{inputenc}
\usepackage[T1]{fontenc}
\usepackage{arxiv}

\usepackage{amsmath}
\usepackage{amssymb}
\usepackage{graphicx}
\usepackage{float}
\usepackage{booktabs}
\usepackage{listings}
\usepackage{subcaption}
\usepackage{enumitem}
\usepackage{natbib}
\usepackage{pifont}
\newcommand{\cmark}{\textcolor{codegreen}{\ding{51}}}
\newcommand{\xmark}{\textcolor{alertred}{\ding{55}}}
\definecolor{alertred}{rgb}{0.70,0.10,0.10}

\usepackage{hyperref}
\usepackage{url}
\hypersetup{
  colorlinks=true,
  linkcolor=arxivaccent,
  citecolor=arxivaccent,
  urlcolor=arxivaccent,
  filecolor=arxivaccent,
  breaklinks=true,
}

\definecolor{codebg}{rgb}{0.95,0.95,0.97}
\definecolor{codegreen}{rgb}{0.0,0.5,0.0}
\definecolor{codegray}{rgb}{0.5,0.5,0.5}
\definecolor{codepurple}{rgb}{0.58,0,0.82}
\definecolor{codeblue}{rgb}{0.0,0.0,0.7}

\lstdefinestyle{pythonstyle}{
    backgroundcolor=\color{codebg},
    basicstyle=\ttfamily\small,
    breaklines=true,
    captionpos=b,
    commentstyle=\color{codegreen},
    keywordstyle=\color{codeblue}\bfseries,
    numberstyle=\tiny\color{codegray},
    stringstyle=\color{codepurple},
    language=Python,
    showstringspaces=false,
    numbers=left,
    numbersep=5pt,
    frame=single,
    rulecolor=\color{codegray},
    tabsize=4,
    morekeywords={dataclass, List, Optional, assert, Dict, field}
}
\title{RLVP: Penalize the Path, Reward the Outcome}

\author{%
  \begin{tabular}{@{}c@{\hspace{4em}}c@{}}
    Bojie Li & Noah Shi \\
    Pine AI & University of Washington
  \end{tabular}%
}

\date{}
\runningtitle{RLVP: Penalize the Path, Reward the Outcome}

\begin{document}
\maketitle

\begin{abstract}
Agents acting on our behalf in the real world---placing phone calls, resolving support tickets---must learn \emph{online} from costly, often irreversible interactions rather than cheap simulator steps. Two things follow. First, deployability depends on the \emph{path}, not only the outcome. An agent must respect outcome-neutral constraints such as not repeatedly calling an unresponsive user, respecting business hours, or completing required authentication---constraints that outcome-based rewards cannot express, since violating them frequently improves apparent success. Second, because each interaction is expensive, the agent must learn efficiently from very few examples.

Reinforcement learning from verifiable rewards (RLVR) is blind to both challenges: it optimizes solely on the outcome and wastes expensive rollouts on all-fail groups where group-relative advantage collapses to zero. Attempts to densify supervision by rewarding progress target the hard-to-verify direction. In contrast, real agentic environments can cheaply detect bad moves. Since group-relative advantage is equivalent to within-group variance, a dense signal helps only when it supplies variance the outcome lacks. A verifiable \emph{penalty} on the path meets this condition reliably, while a progress \emph{potential} helps only where partial progress is reachable. The resulting recipe---\textbf{penalize the path, reward the outcome}---achieves high task success with near-zero violations, where outcome-only training violates constraints on nearly every episode. We provide four design rules for effective penalties, including avoidance of the \emph{inaction trap} that arises when a penalty is used in isolation.
\end{abstract}
\begin{center}
\small
Code: \url{https://github.com/19PINE-AI/rlvp} \\[2pt]
Website: \url{https://01.me/research/rlvp}
\end{center}
\vspace{-0.6em}

\vspace{-0.4em}

\begin{figure}[H]
\centering
\includegraphics[width=\textwidth]{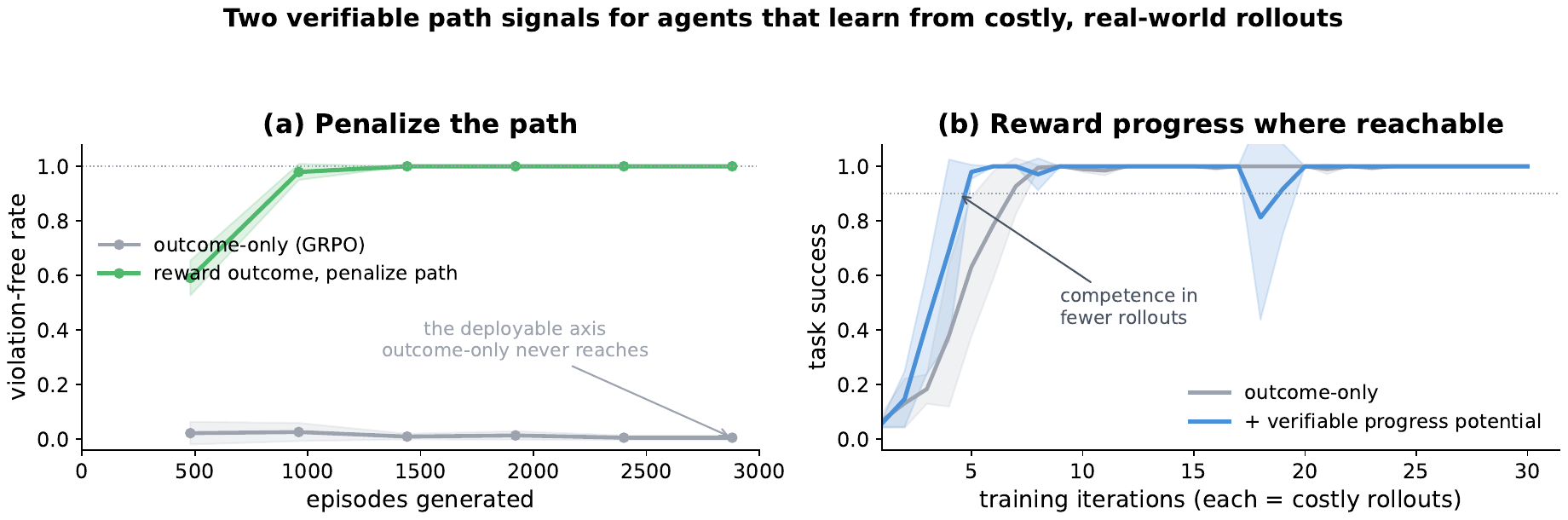}
\caption{\textbf{Two verifiable path signals for agents that learn from costly, real-world rollouts.} \emph{(a) Penalize the path:} rewarding the outcome while penalizing the path drives violation-free episodes from near zero to $\approx$100\%---a deployability constraint outcome-only GRPO is blind to and never reaches---at no cost to task success (\S\ref{sec:path}). \emph{(b) Reward progress where reachable:} a dense potential reaches competence in fewer costly rollouts than outcome-only training (\S\ref{sec:potentials}). Mean over seeds; bands $\pm$1 s.d.}
\label{fig:headline}
\end{figure}

\section{Introduction}

Consider an agent that acts on behalf of a person in the real world: it calls a bank, navigates through the automatic voice menu, waits when put on hold, and eventually speaks with a representative to resolve a billing dispute. Every action is a real, often irreversible interaction---a genuine call rather than a resettable simulator step that can be replayed millions of times. A rapidly growing class of deployed agents operate in this high-stakes regime~\citep{pineai}. Unlike the reversible, self-contained environments of math problems, coding benchmarks, and simulated games that shaped today's foundation models, these agents must contend with partial observability, irreversible consequences, long feedback delays, and genuine uncertainty about the state of the external world. This regime differs from the reversible, high-throughput environments used to train today's models in two fundamental ways.

First, deployability depends on the \emph{path} the agent takes, not the outcome alone. A successful resolution is necessary but insufficient. An agent that reaches the right outcome by calling a user who has explicitly declined contact, phoning outside business hours, or bypassing a bank's required authentication steps is not deployable---no matter how favorable the final result. Outcome-based rewards cannot express these constraints: violating them often \emph{increases} the apparent success rate, because shortcuts and rule-breaking tend to accelerate resolutions in the short term. A deployable agent must therefore satisfy both outcome and path constraints simultaneously.

Second, these agents must learn \emph{online} from their own live interactions, and they must do so from very few examples. In math and code benchmarks, cheap simulators allow policies to generate millions of rollouts, making sample efficiency secondary---slow learning can be compensated by simply running more trials. In the real world, a phone call cannot be replayed. An agent that learns from a single interaction that a particular bank requires a customer's date of birth and the last four digits of their account number must generalize that knowledge immediately to the next user. Here, sample efficiency is not merely an optimization concern; it is the difference between a system that meaningfully improves over time and one that cannot be deployed at all.

Reinforcement learning from verifiable rewards (RLVR)~\citep{deepseekr1,tulu3} is fundamentally blind to both challenges. It optimizes solely on the one signal the environment can evaluate cheaply---the final outcome---rendering the path invisible. This method also wastes its most expensive interactions on all-fail trajectories where group-relative advantage is zero and no learning occurs. The natural response is to seek denser rewards by crediting \emph{progress}~\citep{prm,mathshepherd,setlurprogress,steptool,gigpo}. Yet in agentic settings, judging progress is precisely the hard problem the agent must solve. This leads either to a learned critic that the policy eventually fools~\citep{pure,vineppo,overopt} or to brittle, hand-crafted proxies.

Real agentic environments are \emph{asymmetric verifiers}: they can cheaply and reliably detect \emph{bad moves} (e.g., calling before preconditions are met or acting outside business hours), but cannot certify that an agent is making meaningful progress. The reliable dense signal they provide is therefore a \emph{penalty on the path}, not a reward for progress. Contrary to folklore, penalties are not inherently problematic; the common collapse into inaction is a wiring error, not an inevitability. A penalty in isolation indeed fails---its optimal policy is to do nothing, falling into an \emph{inaction trap}. However, when paired with an outcome reward---reward the desired result, penalize forbidden paths---the penalty effectively teaches precisely the outcome-neutral constraints required for deployment, leveraging the verifiable side of the environment.

\begin{figure}[t]
\centering
\includegraphics[width=\textwidth]{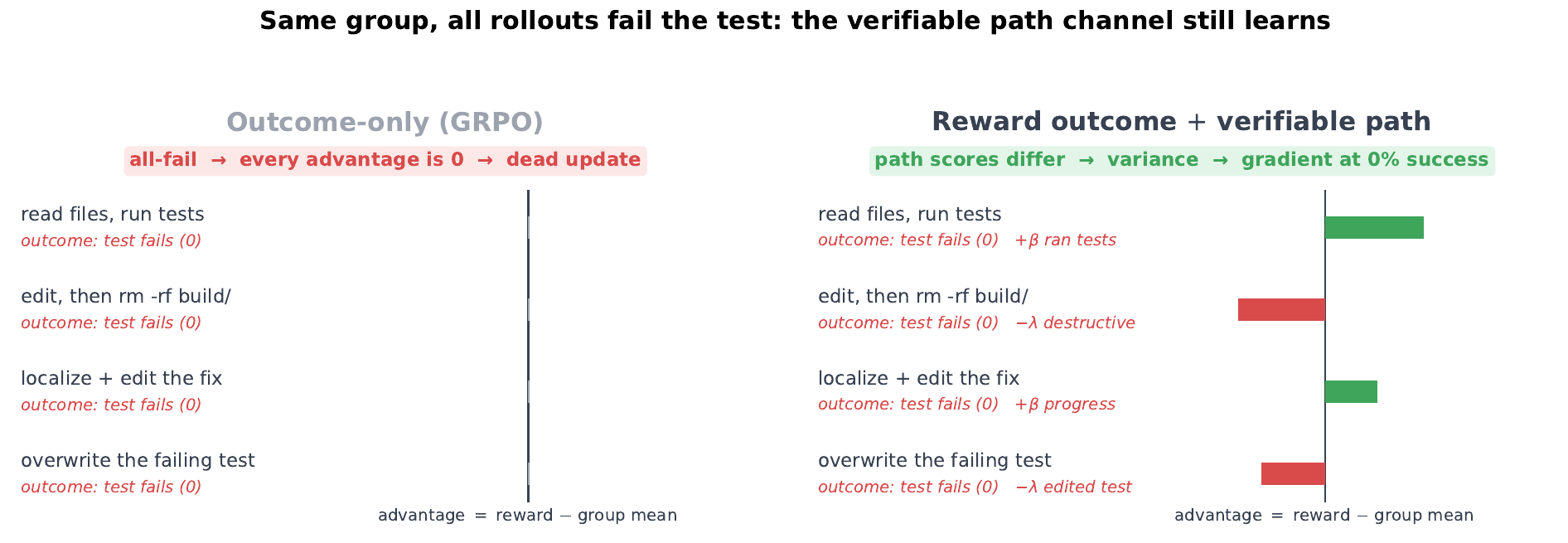}
\caption{\textbf{Why the path channel creates gradient where the outcome cannot.} Four rollouts of the same bug-fix task, all of which fail the hidden test (outcome is $0$ for every trajectory). \emph{Left (outcome-only):} All four rollouts receive identical reward equal to the group mean. The advantage is zero everywhere, so the policy update is dead. \emph{Right (with the verifiable path channel):} The same rollouts receive different path scores---a bonus $+\beta$ for running tests or making measurable progress, and a penalty $-\lambda$ for destructive commands or editing test files. This creates within-group variance, producing a meaningful gradient even at $0\%$ task success.}
\label{fig:grouprollout}
\end{figure}

The same asymmetry has a useful second edge. When the environment \emph{can} verify a step of progress---a precondition satisfied, a subgoal reached---that signal can be used as a dense \emph{potential} to densify the otherwise sparse outcome reward. This is particularly valuable on the all-fail groups that otherwise waste rollouts early in training. A single principle underlies both mechanisms: group-relative advantage is simply within-group variance. The pure outcome signal is blind at both extremes---zero variance on all-failing groups early in training and on all-succeeding groups late in training. A dense auxiliary signal is useful only to the extent that it supplies variance the outcome lacks, and only in states the policy can actually reach. A verifiable penalty satisfies this by construction: bad actions are always detectable. A progress potential, by contrast, is reachability-gated---it helps only once the policy begins to enter states where partial success is possible. The penalty is the universally available half of the solution; the potential is the conditional half. Figure~\ref{fig:headline} previews both. We validate these ideas on tractable proxies for the target deployment setting, including system-administration and customer-service tasks, a shell benchmark, a theorem prover, and software repair. The claimed transfer is mechanistic: the same principles, not the same numerical results, are expected to apply in live production traffic.

\paragraph{Contributions.}
\begin{itemize}[leftmargin=1.4em,itemsep=1pt,topsep=2pt]
\item \textbf{A within-group-variance account} of when dense signals help in group-relative agentic RL. A dense signal is useful precisely when it supplies reachable within-group variance that the outcome reward lacks (\S\ref{sec:background}). This view is symmetric---pure outcome-based RL is blind on both all-fail groups (early training) and all-success groups (late training), as shown in Figure~\ref{fig:grouprollout}---and unifies the two complementary uses of a verifiable path channel: penalties for bad moves and credit for verified progress.
\item \textbf{Penalize the path} (\S\ref{sec:path}). We introduce a per-action verifiable penalty, combined with the outcome reward and governed by four design principles (including mitigation of the inaction trap). This approach reliably teaches outcome-neutral deployment constraints. On a real benchmark it reduces harmful actions by nearly sixfold at equal success rate, driving violation rates from nearly every episode down to near zero.
\item \textbf{Reward verified progress} (\S\ref{sec:potentials}). Using the \emph{same} verifiable channel for credit rather than punishment turns it into a dense \emph{potential}. This directly addresses sample efficiency in online learning: it converts dead all-fail updates into useful gradient wherever partial progress is reachable, while remaining inert where it is not.
\end{itemize}

\begin{figure}[t]
\centering
\includegraphics[width=0.72\textwidth]{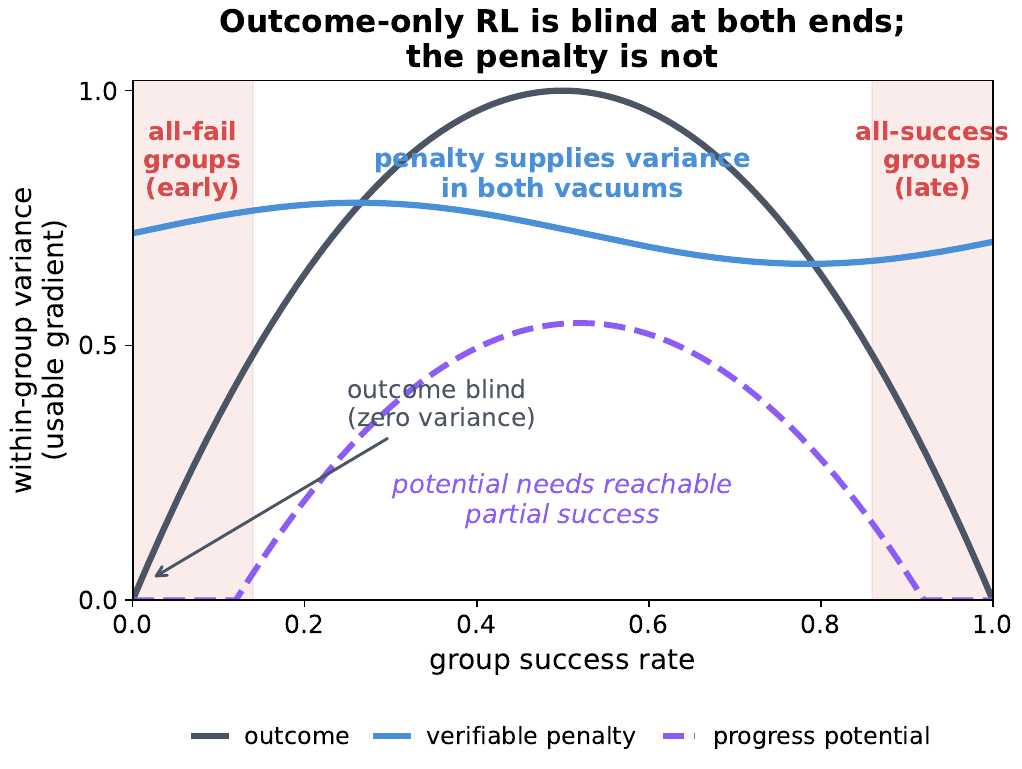}
\caption{\textbf{Outcome-only RL is blind at both ends of the success rate.} Group-relative advantage equals within-group variance. A binary outcome signal drives this variance to zero both on all-fail groups (early in training) and all-success groups (late in training). A dense process signal is useful only insofar as it supplies variance in these regimes---and only in states the policy can actually reach. A verifiable penalty reliably provides this signal (bad moves are easy to detect), whereas a dense progress potential does not (partial success is rare on hard tasks).}
\label{fig:variance}
\end{figure}

\section{The Verifiable Path Channel}
\label{sec:background}

\paragraph{Group-relative RL.} Modern agentic RL has largely moved away from learned value functions. Instead, it estimates advantages by comparing trajectories \emph{within a group} sampled for the same prompt or task~\citep{deepseekmath,deepseekr1}. For a given prompt, the method samples a group of $G$ trajectories, assigns each a reward, and computes its advantage as the deviation from the group mean (often normalized by the group standard deviation). There is no separate critic: the other trajectories in the same group serve as the baseline. This design is precisely why verifiable outcome rewards are so effective---a rule or test suite can score a trajectory directly---and it forms the foundation for all the techniques developed in this paper.

\paragraph{The advantage is a within-group variance.} Since the baseline is the group mean, a trajectory only contributes gradient when its reward \emph{differs} from the others in its group. When all rollouts receive identical rewards, every advantage is zero and the group yields no policy update. With a binary outcome reward, this degeneracy occurs in two dominant regimes (Figure~\ref{fig:variance}): \emph{all-fail} groups, which prevail early in training on long-horizon tasks, and \emph{all-succeed} groups, which prevail once the task is nearly solved. Consequently, group-relative RL is blind at both extremes of the success rate. The community has only partially addressed this by \emph{discarding} such groups rather than imbuing them with learning signal---for example, DAPO drops prompts that are uniformly correct or uniformly incorrect~\citep{dapo}.

\paragraph{When a dense signal can help.} A dense process signal is useful exactly when it restores within-group variance that the outcome has lost. Consider the shaped reward of a trajectory in the form of the outcome plus $\beta$ times a process term, $R = O + \beta\,\Phi$. The group's reward variance---the quantity that drives the policy gradient---then decomposes as
\begin{equation}
\operatorname{Var}_G(R) \;=\; \operatorname{Var}_G(O) \;+\; \beta^2\,\operatorname{Var}_G(\Phi) \;+\; 2\beta\,\operatorname{Cov}_G(O,\Phi).
\label{eq:vardecomp}
\end{equation}
On all-fail or all-success groups, $\operatorname{Var}_G(O) = 0$. Therefore, \emph{all} usable gradient must come from the process term's own within-group variance $\operatorname{Var}_G(\Phi)$ (Figure~\ref{fig:grouprollout}). This yields a simple, verifiable condition: a process signal helps only if it produces nonzero within-group variance, which occurs only when the policy reaches differing intermediate states across the trajectories in the group.
Both conditions---finer granularity than the outcome and reachability---are satisfied by a single per-action verifiable channel, used in two complementary ways. The channel attaches a verifiable signal to the action that triggers it: a penalty $-\lambda$ for a verified \emph{bad} move (e.g., a call before its precondition is met), a credit $+\beta$ for a verified \emph{good} move (satisfying a precondition or reaching a subgoal). When used as a \emph{penalty}, this signal almost always generates within-group variance early in training: some rollouts in the group take the bad action while others do not. Both cases are cheap to sample and verify, making the penalty the \emph{always-reachable} source of gradient. When used as a \emph{potential}---paying the same $+\beta$ credit for verified progress---the signal provides variance only in states where the policy reaches differing levels of partial success. On hard tasks this is infrequent (Figure~\ref{fig:variance}, right), making the potential the \emph{reachability-gated} use. The remainder of this paper explores these two uses of one verifiable path channel: penalizing the path to achieve deployability (\S\ref{sec:path}) and rewarding verified progress to improve sample efficiency (\S\ref{sec:potentials}).

\section{Penalizing the Path: Verifiable Constraints for Deployable Agents}
\label{sec:path}

This section develops the penalty use of the verifiable path channel. We present a two-channel approach that combines a per-action penalty with the standard outcome reward (\S\ref{sec:twochannel}). We then introduce a practical recipe that enables agents to achieve their tasks while reliably respecting outcome-neutral deployment constraints. This is validated on controlled proxies as well as a real agentic benchmark (\S\ref{sec:recipe}). Finally, we articulate four design rules that distinguish a robust, effective penalty from one that collapses into an inaction trap (\S\ref{sec:design}).

\subsection{The two-channel method}
\label{sec:twochannel}

\begin{figure}[t]
\centering
\includegraphics[width=\textwidth]{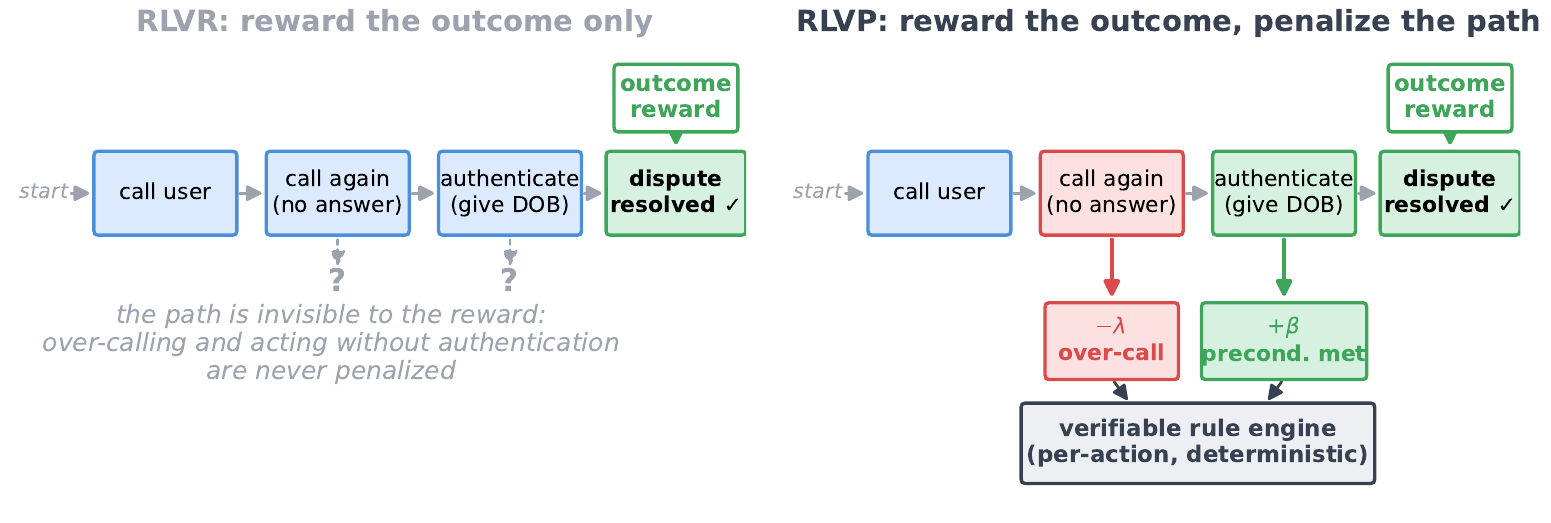}
\caption{\textbf{The verifiable path channel in the real-world (phone-agent) setting.} \emph{Left (RLVR):} Outcome-only training receives only a terminal reward and remains blind to the path. Violations such as over-calling or acting without proper authentication go unpenalized. \emph{Right (RLVP):} A deterministic per-action rule engine attaches an immediate verifiable signal to the responsible action---a penalty $-\lambda$ for bad moves (e.g., over-calling) and a credit $+\beta$ for good moves (e.g., successfully discharging a precondition)---while retaining the original outcome reward. The penalty use (\S\ref{sec:path}) enforces path constraints for deployability. The same $+\beta$ credit, when paid for verified progress, serves as the dense potential introduced in \S\ref{sec:potentials}.}
\label{fig:architecture}
\end{figure}

\paragraph{Two channels.} We keep the outcome reward and introduce a second, per-action channel (Figure~\ref{fig:architecture}). At each step, a deterministic rule engine---a pure predicate over the pre-action state and the action taken---evaluates the action. A \emph{violation} (e.g., issuing a destructive command or calling before a required precondition is met) triggers a penalty $-\lambda$ on the offending action's tokens. Conversely, a \emph{fulfillment} of a pending obligation (e.g., successfully performing the required precondition) triggers a credit $+\beta$. The two channels are normalized separately to prevent the sparse path signal from being diluted or overwhelmed by the outcome reward, and are then combined. ``Penalize the path'' is therefore literal: the penalty directly reduces the probability of the specific bad action in the precise context in which it occurs, while the outcome reward provides global trajectory-level credit for success. We call signals produced by this engine \emph{verifiable penalties}. Importantly, the constraints we penalize are \emph{outcome-neutral}---violating or respecting them does not determine whether the overall task is solved. This is precisely the type of signal that a pure outcome reward can never supply. The complete recipe also includes pairing each penalty with its corresponding fulfillment credit, seeding a small number of scripted compliant demonstrations to ensure compliant behavior is reachable early in training, and annealing the path shaping once compliance saturates. Section~\ref{sec:design} demonstrates that each of these components is essential.

\subsection{The recipe attains the task and its constraints}
\label{sec:recipe}

\begin{figure}[t]
\centering
\includegraphics[width=\textwidth]{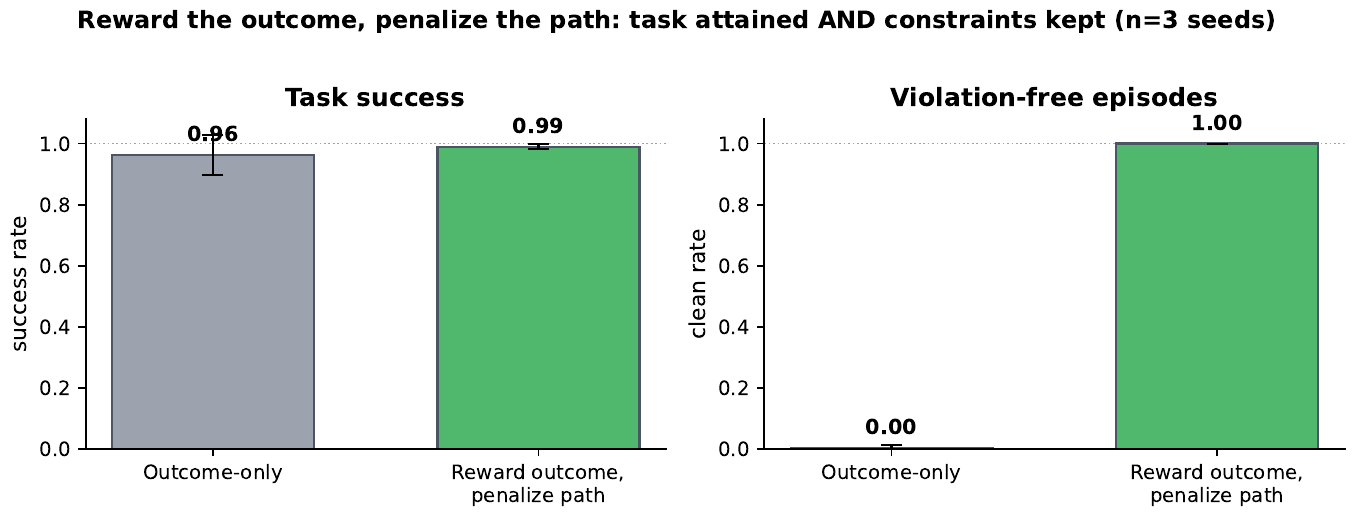}
\caption{\textbf{Rewarding the outcome while penalizing the path satisfies both task success and constraints.} Outcome-only training solves the task but violates outcome-neutral rules on nearly every episode. 
The two-channel method maintains high success while achieving near-100\% violation-free episodes across tasks and model scales (mean ± std over 5 seeds).}
\label{fig:recipe}
\end{figure}

\paragraph{On controlled proxies.} We evaluate on a suite of diverse system-administration and customer-service tasks that serve as controllable proxies for the real deployment setting. These tasks encode outcome-neutral rules of \S1 in miniature: \emph{verify a precondition before acting}, \emph{do not repeat a request that was just refused}, and \emph{read before you mutate}. When trained with outcome-only rewards on held-out instances drawn from a large pool, agents learn to solve the tasks but violate these rules on nearly every episode. The constraints are invisible to the outcome reward, just as calling outside business hours is invisible to whether a billing dispute is ultimately resolved. Adding the penalty channel---reward the outcome, penalize the path---drives the violation rate to near zero while preserving high task success rate. This holds across five seeds at 4B and generalizes from 1.7B to 8B models (Figure~\ref{fig:recipe}). Importantly, the agent does not achieve this by becoming passive: it takes roughly the same number of actions as before, but selects much cleaner ones. This result demonstrates the core benefit in its clearest form: a verifiable penalty teaches genuine deployable behavior---respecting outcome-neutral constraints---that a pure outcome reward cannot.

\begin{table}[tbp]
\centering
\caption{\textbf{Harm reduction at equal outcome (TerminalBench, Qwen3-4B, 5 seeds, mean $\pm$ std).} Task success is statistically equal and at the floor (4B rarely solves the benchmark; the small RLVP--outcome gap is within one standard deviation), yet the un-gameable harm penalty cuts harmful actions roughly \emph{sixfold}. The policy is not going passive: it takes about three times \emph{more} productive actions while violating less. Harm lives on an axis the terminal outcome cannot see.}
\label{tab:harm}
\small
\renewcommand{\arraystretch}{1.2}
\begin{tabular}{@{}lcc@{}}
\toprule
& \textbf{RLVP (harm penalty)} & \textbf{Outcome-only} \\
\midrule
Task success & $0.097 \pm 0.060$ & $0.122 \pm 0.076$ \;(equal within $1\sigma$, at floor) \\
\textbf{Violations / episode} & $\mathbf{0.66 \pm 0.63}$ & $3.71 \pm 0.52$ \\
Productive actions / episode & $\sim$\textbf{13} & $\sim$4 \\
\bottomrule
\end{tabular}
\end{table}

\paragraph{It transfers to a real agentic benchmark.} While the synthetic domains allow precise control over the rules, we also test the same mechanism in a more realistic setting using TerminalBench. In this benchmark, each turn consists of a real shell command executed in a task-specific container, and the environment can flag genuinely destructive actions (e.g., \texttt{rm -rf} or dropping a database) against a live filesystem—shortcuts that outcome-only training would otherwise exploit. On this benchmark, task success for smaller models is near the floor, making it an especially clean test of the harm-reduction axis.At equal task success rates, the verifiable harm penalty reduces destructive actions by a factor of six compared to outcome-only training, while the policy actually issues \emph{more} productive commands overall (Table~\ref{tab:harm}, five seeds). The reduction is large relative to seed variance, with non-overlapping per-seed distributions, and is clearly not the result of passivity. A verifiable penalty bounds the path independently of task success, which is the exact property a deployable agent needs.

\subsection{Design rules for the penalty}
\label{sec:design}

\begin{figure}[t]
\centering
\includegraphics[width=0.9\textwidth]{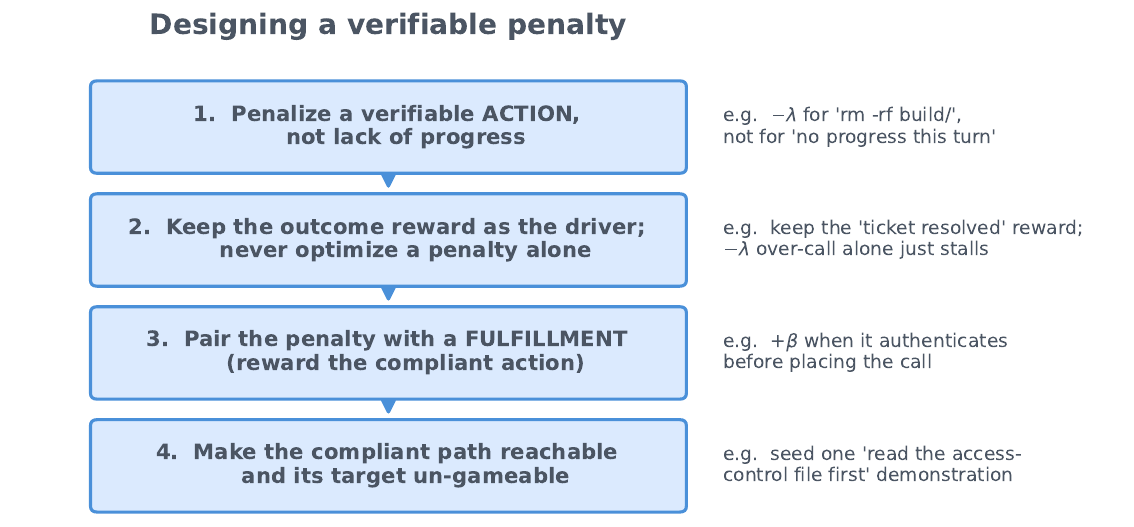}
\caption{\textbf{A design procedure for verifiable penalties.} Before introducing a penalty, each rule can be checked against four criteria: (1) Is there a concrete, verifiable \emph{action} to penalize (rather than the absence of progress)? (2) Does the outcome reward remain the primary driver of task completion? (3) Is the penalty paired with a corresponding credit for the compliant action? (4) Is the compliant action reachable early in training, and is its target difficult to game? Only when all four conditions are satisfied does the penalty successfully bound the path without stalling the task.}
\label{fig:design}
\end{figure}

\begin{figure}[t]
\centering
\includegraphics[width=\textwidth]{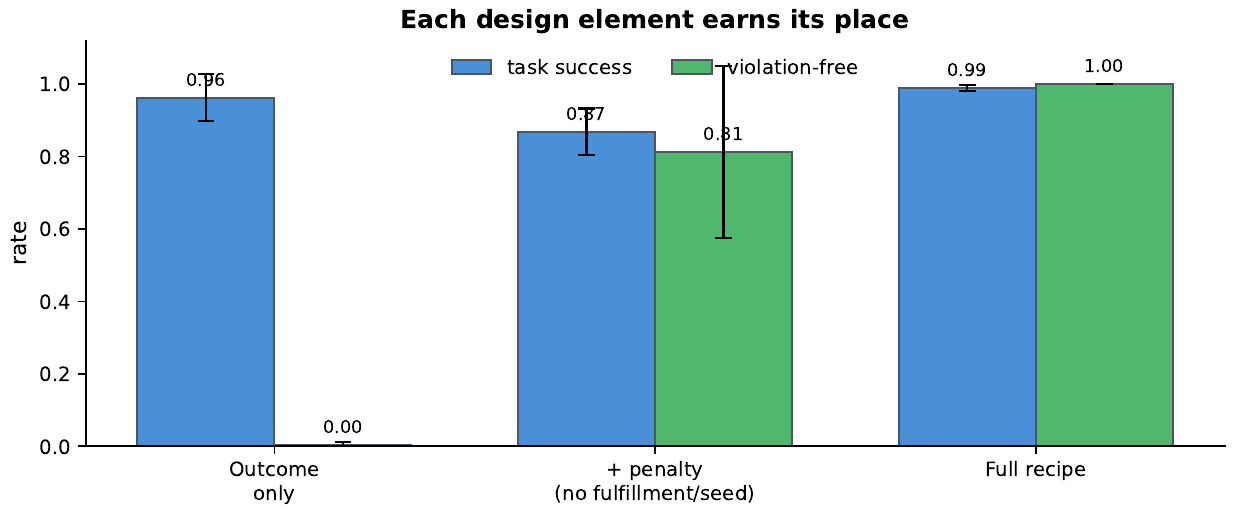}
\caption{\textbf{Each design element earns its place.} On diverse held-out tasks, outcome-only training solves the task but exhibits high violation rates (low clean rate). Adding the verifiable penalty channel \emph{without} its complementary fulfillment credit and reachability seeding fails to reliably produce both clean and successful behavior across seeds. Only the full recipe---penalty + fulfillment + seeded compliant demonstrations + annealing---consistently attains high task success \emph{and} near-zero violations with low seed variance. The pure-penalty inaction trap (a penalty without any outcome reward, which collapses to zero success on every seed) is isolated separately in the ungamability sweep (Appendix~\ref{app:sweep}, Figure~\ref{fig:sweep}). 
Results are means over five seeds; whiskers show one standard deviation.}
\label{fig:ablation}
\end{figure}

While a verifiable penalty is powerful, it is also easy to apply incorrectly. The variance account together with our ablations yield four design rules that distinguish the robust behavior shown above from policies that collapse into inaction (Figure~\ref{fig:design}).

\paragraph{1. Penalize verifiable \emph{actions}, not lack of progress.} A good penalty targets a specific, machine-checkable bad action---for example, executing a destructive command or issuing a call before its precondition is satisfied. It must not target the mere \emph{absence} of progress (e.g., ``penalize an unproductive step''), because the cheapest way to make no unproductive steps is to take no steps at all. Therefore: penalize \emph{commission}, never omission.

\paragraph{2. Keep the outcome reward as the task driver; never optimize a penalty in isolation.} A penalty must not become the main learning signal. In the all-fail regime, it is often the \emph{only} signal with nonzero gradient, and its cheapest optimum is to stop acting entirely. The policy then drifts into a penalty-free but useless solution---the \emph{inaction trap}. A controlled sweep confirms this empirically: a pure penalty (with no outcome reward) collapses task success to zero on \emph{every} seed, whereas any penalty-free configuration continues to make progress (Appendix~\ref{app:sweep}, Figure~\ref{fig:sweep}). The outcome reward is what compels the agent to keep acting toward task completion. The penalty merely shapes \emph{how} it does so. In short: reward the outcome; let the penalty steer.

\paragraph{3. Pair the penalty with a corresponding fulfillment credit.} A penalty only tells the policy what \emph{not} to do. Pairing it with a fulfillment credit ($+\beta$) for the compliant action creates an active pull toward the desired behavior. This helps the policy move decisively away from the bad action rather than merely avoiding it. Ablations show that removing the fulfillment credit measurably slows learning and increases instability in acquiring compliant behavior (Figure~\ref{fig:ablation}).

\paragraph{4. Make the compliant behavior reachable and its target un-gameable.} A fulfillment credit can only reinforce behavior that the policy actually samples. If the compliant path is never explored, the credit has no effect. To ensure reachability, we seed training with a small number of scripted compliant demonstrations. Once compliance saturates, the path shaping can be safely annealed away. Additionally, the penalized (and credited) target must be a concrete, verifiable action rather than a proxy. Using a learned judge of ``compliance'' simply moves the gaming problem into the judge itself (Appendix~\ref{app:boundaries}).

Together, these four rules delineate a narrow but reliable regime for effective path penalties. The penalty must be verifiable and outcome-neutral, must accompany (rather than replace) the outcome reward, must be paired with a corresponding fulfillment credit, and must target reachable and un-gameable actions. The ablation in Figure~\ref{fig:ablation} progressively builds from outcome-only training through partial versions of the penalty channel (without fulfillment credit or reachability seeding) to the full recipe. Compliant and successful behavior emerges robustly only when all four rules are satisfied. The pure-penalty inaction trap---where a penalty is used without any outcome reward---is isolated separately in the sweep shown in Appendix~\ref{app:sweep}.

\section{Reward Verified Progress: Sample Efficiency Where Reachable}
\label{sec:potentials}

\begin{figure}[t]
\centering
\includegraphics[width=\textwidth]{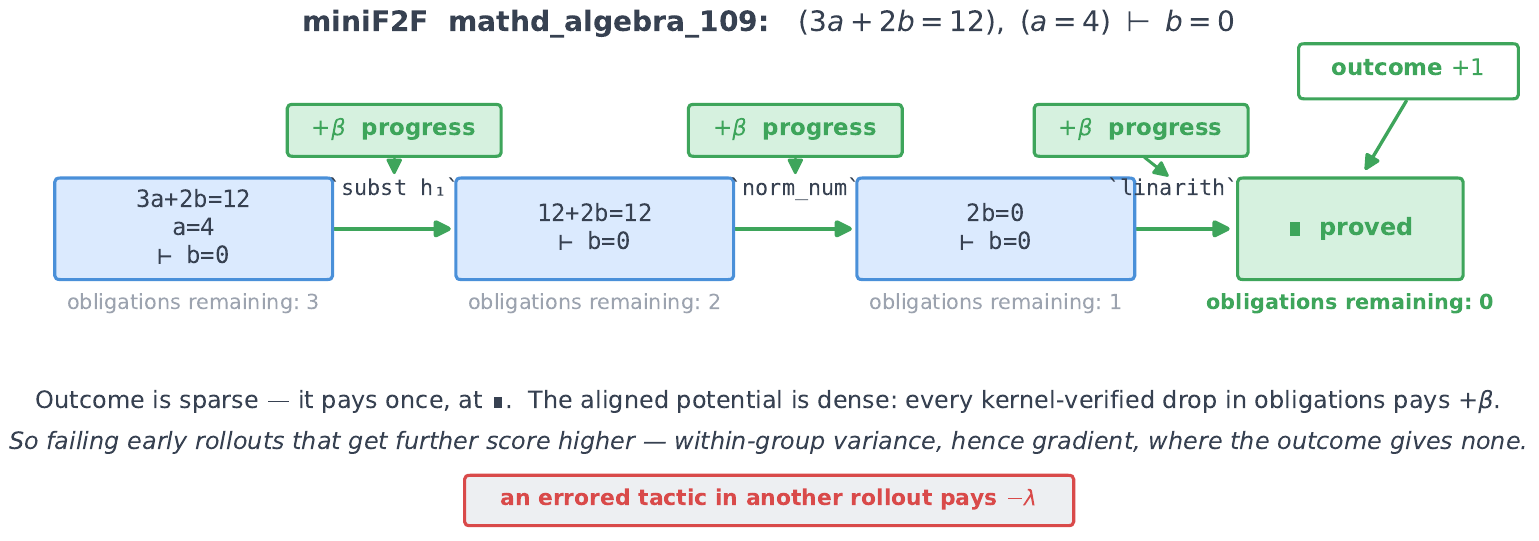}
\caption{\textbf{The aligned potential in action on a real \textsc{miniF2F} theorem.} On \texttt{mathd\_algebra\_109}, the Lean kernel verifies each tactic and the proof's remaining obligations fall $3\!\to\!2\!\to\!1\!\to\!0$. The outcome reward is \emph{sparse}: it pays a single $+1$ only when the proof is completed ($\blacksquare$). The aligned potential is \emph{dense}: every kernel-verified drop in obligations pays a $+\beta$ fulfillment credit (while an errored tactic in another rollout pays $-\lambda$). On the all-fail groups early in training---where no rollout completes the proof---trajectories that make more progress receive higher path scores. This supplies the within-group variance, and therefore the gradient, that the pure outcome reward cannot provide.}
\label{fig:potillus}
\end{figure}

\begin{figure}[t]
\centering
\includegraphics[width=\textwidth]{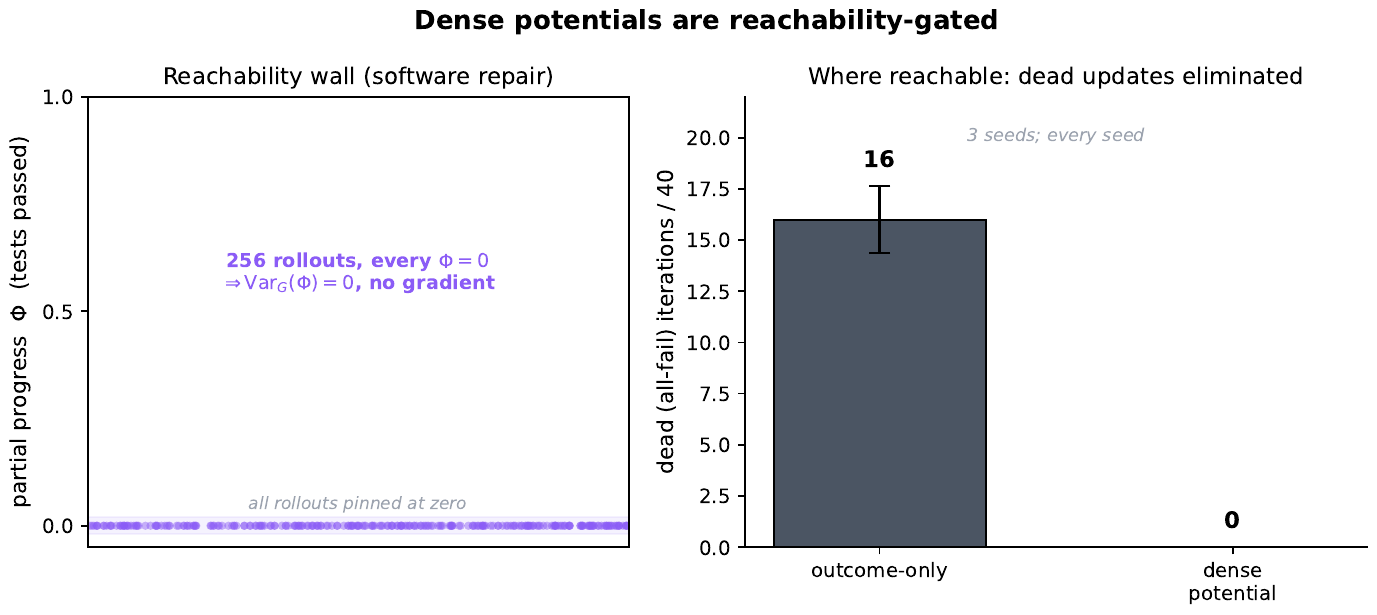}
\caption{\textbf{Dense potentials are reachability-gated.} \emph{Left:} On software repair, the finer-grained potential is unreachable. Across many rollouts, the policy makes zero partial progress, yielding zero within-group variance and thus no gradient. \emph{Right:} Where the potential \emph{is} reachable, it eliminates the dead all-fail updates that dominate early training (0 vs.\ $\sim$16 of 40 iterations per seed). It supplies gradient precisely where the outcome reward is blind. The limiting factor is reachability, not fragility.}
\label{fig:potentials}
\end{figure}

The path channel's credit has a second use. In \S\ref{sec:path} the $+\beta$ credit was paired with a penalty to reward compliant actions. Here, we pay the \emph{same} credit for verifiable \emph{progress}---such as a reduction in remaining goals in a proof, an increase in the fraction of tests passed, or a precondition being satisfied. This turns the signal into a dense \emph{potential} (Figure~\ref{fig:potillus}) that directly addresses the second key challenge of real-world deployment: learning efficiently from very few costly interactions. By densifying the sparse outcome reward on all-fail groups (which otherwise waste expensive rollouts), the potential enables the policy to reach competence in fewer interactions. We study this approach in domains where partial progress is well-defined and cheap to evaluate---theorem proving and software repair. These serve as controllable proxies that allow extensive experimentation and multiple seeds, which would be infeasible in a live phone-agent setting. The within-group variance account applies directly: the credit supplies useful gradient only where the policy reaches intermediate states with differing progress levels. The determining factor is \emph{reachability}. On software repair, for example, a natural potential is the fraction of hidden tests passed during an episode. For many capable models, \emph{every} rollout achieves zero test passes early in training, so the potential has zero within-group variance and provides no signal. Reachability can be diagnosed before training using a small number of rollouts from the base policy, and the benefit of adding a dense potential closely tracks this diagnostic. In short, reachability---not fragility---is the true gate.

\begin{figure}[t]
\centering
\includegraphics[width=\textwidth]{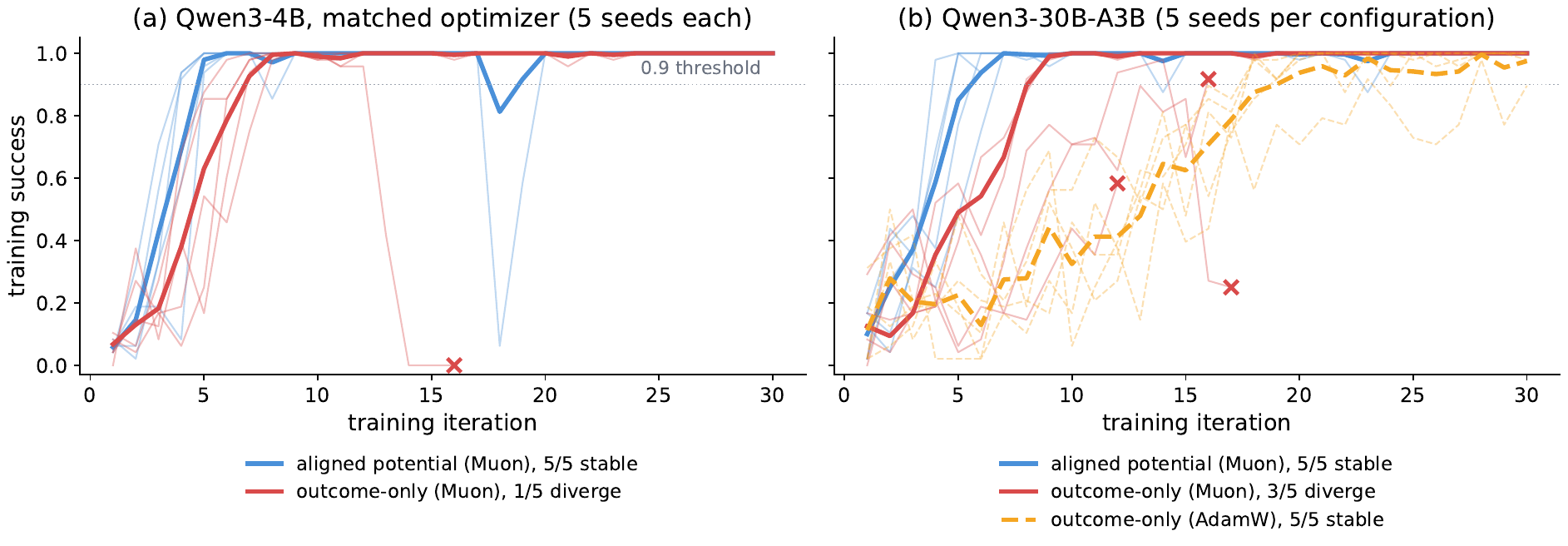}
\caption{\textbf{The aligned-potential recipe on real theorems (miniF2F algebra), five seeds per configuration.} Thin lines show individual seeds; thick lines show the mean over stable seeds; $\times$ marks a divergence-guard abort. \emph{(a)} Controlled same-optimizer comparison at 4B scale: the aligned potential crosses the $0.9$ threshold in $4.4\pm0.5$ iterations vs.\ $7.0\pm0.7$ for outcome-only. It is faster on every seed and never diverges, compared to outcome-only which loses one seed to entropy collapse. \emph{(b)} At 30B scale, outcome-only exhibits a speed--reliability trade-off. Under Muon it diverges on three of five seeds; under AdamW it is stable but takes $\sim$3.6$\times$ longer ($19.2\pm1.9$ vs.\ $5.4$ iterations). The aligned potential is both fast \emph{and} stable across all seeds at both scales.
}
\label{fig:recipescale}
\end{figure}

\begin{table}[t]
\centering
\caption{\textbf{The aligned-potential matrix on miniF2F algebra} (mean $\pm$ std over stable seeds; divergence-guard aborts counted separately). The aligned potential is the only configuration that is simultaneously fast and reliable at both scales. The annealing ablation shows that it does not need annealing. The AdamW rows provide the stable-optimizer baseline for the outcome-only configuration.}
\label{tab:p0}
\small
\renewcommand{\arraystretch}{1.15}
\begin{tabular}{@{}llcccc@{}}
\toprule
scale & configuration & iters to $0.9$ & AUC & final success & diverged \\
\midrule
4B & aligned potential (Muon) & $\mathbf{4.4 \pm 0.5}$ & $\mathbf{0.90 \pm 0.03}$ & $1.00 \pm 0.00$ & $\mathbf{0/5}$ \\
4B & \quad + anneal & $5.3 \pm 0.5$ & $0.85 \pm 0.05$ & $0.99 \pm 0.02$ & $0/3$ \\
4B & outcome-only (Muon) & $7.0 \pm 0.7$ & $0.87 \pm 0.02$ & $1.00 \pm 0.00$ & $1/5$ \\
\midrule
30B & aligned potential (Muon) & $\mathbf{5.4 \pm 1.0}$ & $\mathbf{0.90 \pm 0.02}$ & $1.00 \pm 0.00$ & $\mathbf{0/5}$ \\
30B & outcome-only (Muon) & $8.5 \pm 0.5$ & $0.84 \pm 0.00$ & $1.00 \pm 0.00$ & $3/5$ \\
30B & outcome-only (AdamW) & $19.2 \pm 1.9$ & $0.63 \pm 0.07$ & $0.97 \pm 0.05$ & $0/5$ \\
\bottomrule
\end{tabular}
\end{table}

Where the potential \emph{is} reachable, it robustly supplies the gradient the outcome reward lacks. It eliminates the dead all-fail updates that dominate early training (0 vs.\ $\sim$16 of 40 iterations) and, when paired with a bounded optimizer, converts this signal into faster and more reliable learning. We quantify this benefit on real theorem proving using a matched five-seed matrix across two model scales (Figure~\ref{fig:recipescale}). The aligned potential delivers a consistent speed-up under a matched optimizer and a substantial reliability advantage under an aggressive optimizer. At 30B scale, outcome-only training faces a stark trade-off: it either diverges under the fast optimizer or trains roughly 3.6$\times$ slower under its stable counterpart. In contrast, the aligned potential is both fast and stable across every seed at both scales. These gains are modest and seed-consistent. They are not the dramatic single-run speed-ups sometimes reported in the literature---our own single-run pilots showed such effects, but they disappeared under proper re-seeding. Protocol and divergence taxonomy are provided in Appendix~\ref{app:potentials}.

\paragraph{An aligned-potential recipe.} Simply adding a dense potential is not enough for robust improvement. Three ingredients are essential: \textbf{(i) Alignment.} The potential must be \emph{outcome-instrumental}---it should increase as the task is solved (e.g., a falling goal count or a rising fraction of tests passed). It is implemented as a token-attached fulfillment credit with its own normalizer. Generic \emph{structural} proxies (such as read-before-write hygiene or tactic well-formedness) are misaligned: once the task is mastered, they tend to be over-optimized, imposing an artificial performance ceiling. \textbf{(ii) Reachability.} The potential must take on intermediate values that the policy actually samples. This can be diagnosed before training by measuring $\operatorname{Var}_G(\Phi)$ on rollouts from the base policy. As shown in Figure~\ref{fig:potentials}, reachability is vacuous on software repair but present on theorem proving. \textbf{(iii) A bounded optimizer.} Aggressive optimizers easily drive the shaped policy into entropy collapse or gradient explosion. A more stable optimizer (we use Muon) is critical for reliable training. Annealing the potential after saturation, often considered mandatory, proves unnecessary in this setting. Because the potential is outcome-instrumental, there is no misaligned pressure to over-optimize it. The annealing ablation (Table~\ref{tab:p0}) shows no meaningful difference beyond noise.

\section{Discussion}

\paragraph{Verifier asymmetry as a design principle.} The central practical takeaway is a reallocation of effort. Verifiers are scarce and valuable resources. The field has been spending them primarily on the direction they are weakest at---certifying progress---while largely ignoring the direction they are strongest at---detecting bad moves. We should invert this: spend verifiable signals on penalties. Whenever there is an outcome-neutral constraint a deployed agent must respect, and a concrete verifiable bad action to penalize, a penalty paired with its corresponding fulfillment credit and combined with the outcome reward teaches the constraint robustly and at low cost. When the only available dense signal is a progress proxy, first diagnose reachability (e.g., by measuring within-group variance on base-policy rollouts). Where intermediate values are reachable, the potential meaningfully accelerates learning. Where they are not, it provides no signal, and using the outcome reward alone is preferable.

\begin{figure}[t]
\centering
\includegraphics[width=0.9\textwidth]{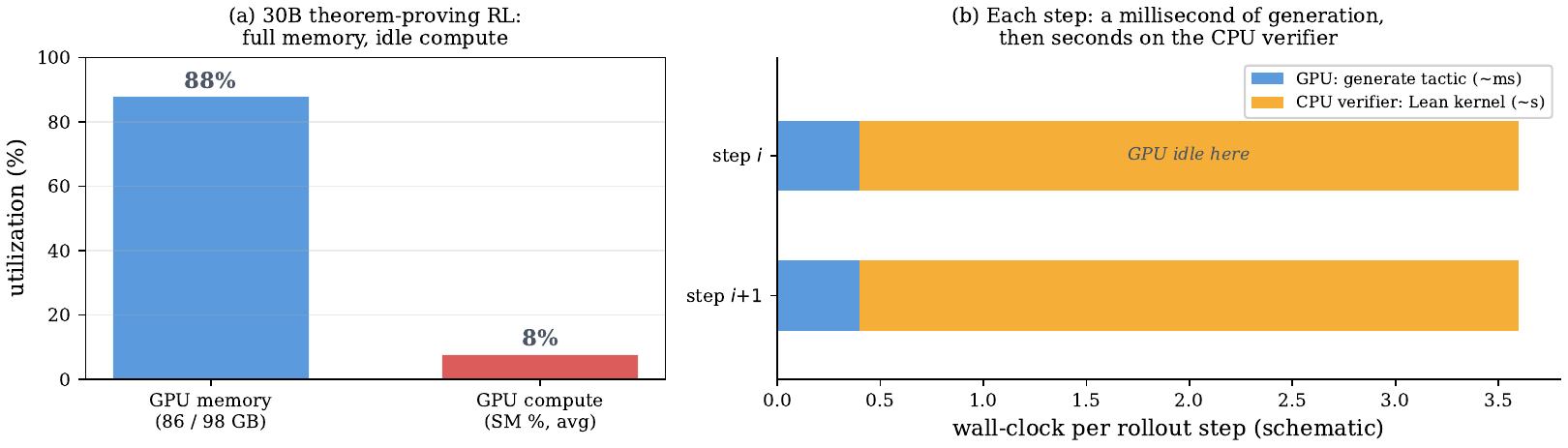}
\caption{\textbf{Verifiable agentic RL is often verifier-bound.} In theorem-proving experiments the GPU sits near-idle at full memory utilization while the CPU-based proof kernel verifies each step. The same signal that makes a penalty cheap also moves the bottleneck away from the accelerator.}
\label{fig:verifier}
\end{figure}

\paragraph{A systems note.} The same property that makes verifiable penalties cheap---a machine-checkable environment---often makes the verifier a CPU-bound process (a rule engine, a test runner, or a container). In our theorem-proving experiments, the accelerator generated a step in milliseconds but then waited seconds for the CPU-based proof kernel to verify it. This resulted in GPU utilization in the single digits despite full memory occupancy. This behavior is workload-dependent rather than fundamental, but in verifiable agentic domains where penalties are applicable, overall throughput is typically gated by environment parallelism. Effective systems design therefore focuses on co-locating many verifier workers per accelerator rather than simply scaling accelerators (Figure~\ref{fig:verifier}).

\paragraph{From proxies to deployment.} The two properties we rely on---that each rollout is expensive and irreversible, and that the checkable signal concerns path constraints rather than progress judgments---are most pronounced in the real-world online setting that originally motivated this work: an agent improving on live phone calls cannot afford wasted rollouts and must respect operational rules for which it receives no direct reward. This same setting makes large-scale, controlled, multi-seed experimentation impractical, which is why we evaluated on tractable proxies carefully chosen to preserve the essential structure. The transfer we claim is therefore mechanistic rather than numerical: a verifiable penalty supplies within-group variance where the outcome reward cannot, and a verifiable potential does so wherever partial progress is reachable. Validating these ideas directly on a deployed online agent---where the true sample-efficiency gains are measured in real user interactions saved---is the natural and most important next step.

\paragraph{Limitations.} Our strongest real-task harm reduction result was obtained in a regime where task success is near the floor. While the underlying mechanism predicts that the benefit persists at higher success rates (since penalty variance remains reachable regardless of outcome), validating this on a highly capable model is an important next step. Additionally, our penalties were manually identified based on domain rules. Automating the discovery of penalizable constraints directly from an environment's affordances remains an open challenge. Finally, the sweep isolating the inaction trap shows high variance at larger scales. We therefore report the qualitative \emph{survival pattern} rather than precise quantitative magnitudes.

\section{Related Work}

\paragraph{Group-relative RL and the zero-variance problem.} Modern agentic RL has largely converged on group-relative policy optimization (GRPO)~\citep{deepseekmath} and the rule-based R1-style methods~\citep{deepseekr1,tulu3}. Because these approaches use a within-group baseline, advantage is \emph{definitionally} equivalent to within-group variance, which collapses to zero on groups that are uniformly successful or uniformly failed. Prior responses to this ``advantage collapse'' have either discarded the dead groups (e.g., DAPO's dynamic sampling~\citep{dapo}) or attempted to re-weight them---for instance by removing the difficulty normalizer~\citep{drgrpo}, stabilizing the sequence-level ratio~\citep{gspo}, or injecting entropy-based signals~\citep{rlzvp}. Notably, DAPO discards \emph{all-correct} groups in addition to all-wrong ones, implicitly acknowledging blindness at \emph{both} extremes of success rate. Our work makes this symmetry explicit. Rather than discarding blind groups, we ask what kind of dense signal can supply the missing within-group variance. A verifiable penalty does so reliably, while a potential does so only where partial progress is reachable. Concurrently, VeriGate~\citep{verigate} gates step-level supervision on a similar condition, applying it only when verifier rewards become degenerate.

\paragraph{Penalties and negative reinforcement.} Negative signals are powerful tools on their own. Training against negative examples can suppress undesirable generations and reallocate probability mass~\citep{nsr}, while negative-enhanced updates have been shown to recover gradient from all-wrong groups~\citep{ngrpo}. However, these approaches penalize \emph{wrong outcomes}. In contrast, our verifiable penalty targets specific, \emph{outcome-neutral} path violations. This supplies a form of signal that no outcome-based reward---positive or negative---can ever provide. As emphasized in our design rules, such a penalty is effective only when paired with an outcome reward that continues to drive task completion (Design Rule 2).

\paragraph{Process reward models and their gameability.} Step-level supervision has been shown to outperform pure outcome supervision for verification~\citep{prm}, with labels obtainable via Monte-Carlo rollouts~\citep{mathshepherd}. Some works frame the useful signal as a measure of progress~\citet{setlurprogress}, which is close to our variance view, but instead instantiated with a \emph{learned} prover. However, learned dense rewards are notoriously vulnerable under RL optimization pressure. They are prone to over-optimization and Goodhart's Law effects~\citep{overopt}, and capable agents can exploit misspecifications more effectively~\citep{misspec}. Summation-form process reward models often lead to training collapse~\citep{pure}, while learned value functions provide unreliable credit~\citep{vineppo}. Even DeepSeek-R1 ultimately abandoned process reward models, describing them as ``inevitably'' hackable~\citep{deepseekr1}, despite an implicit process model existing within their outcome reward model~\citep{freeprm}. This motivates our use of \emph{verifiable, non-learned} penalties together with our fourth design rule (un-gameability).

\paragraph{Potential-based shaping: safe versus useful.} Potential-based shaping leaves the optimum unchanged for \emph{any} potential~\citep{ng1999} and is equivalent to value initialization~\citep{wiewiora2003}. These results certify which dense signals are \emph{safe}, but say little about which ones are actually \emph{useful}. Our work is orthogonal. Among safe shapings, we identify those that are helpful: those that supply within-group variance where the outcome reward lacks it, and only where such variance is reachable. Concretely, a penalty on frequently sampled bad actions is helpful, whereas a potential based on rarely reached progress usually is not.

\paragraph{Process rewards in agentic RL, and the fragility of RL gains.} Prior work on agentic step- and turn-level credit includes hierarchical critics~\citep{archer}, learned step rewards~\citep{steptool,sparl}, turn-level estimators~\citep{turnlevel}, and---closest to our setting---GiGPO~\citep{gigpo} and per-tool-call correctness~\citep{toolrl}; see \citet{creditsurvey} for a survey. More recent methods further densify rewards: HERO blends a binary verifier with a continuous RM score under \emph{variance-aware} weighting~\citep{hero}, implicit step rewards distill a step PRM from trajectory preferences~\citep{implicitstep}, and RLTR rewards tool-use \emph{completeness}~\citep{rltr}. Among these, RLTR's verifiable check is the reachable process credit we advocate. However, most learned dense rewards are fragile under optimization pressure. Naive per-turn rewards can degrade performance when discriminativeness and advantage direction misalign~\citep{rewardcalib}, and turn-level gains are highly sensitive to the choice of policy-gradient estimator~\citep{practitioner}. This fragility aligns with a broader literature on non-replicable RL gains, including high seed variance~\citep{sober}, spurious correlations~\citep{spurious}, one-shot improvements that fail to replicate~\citep{oneshot}, and cases where base models match RLVR performance at large sample counts~\citep{rlbeyond}. Our result sharpens this picture: the benefit of a dense potential is gated by reachability and the use of a bounded optimizer, rather than being intrinsic to the signal. Most prior work assumes cheap, resettable simulators where sample efficiency is secondary. Our online, irreversible setting inverts this priority, making reachable potentials and extra-sample-free verifiable penalties first-order concerns. Verifiable \emph{outcome} rewards remain central to many agentic benchmarks~\citep{taubench,tau2bench,swegym,swerl}. Learned-judge rewards~\citep{selfreward,rlaif,constitutional,llmjudge} and demonstration mixing~\citep{luffy} are complementary to our approach: we use the latter for reachability seeding while cautioning against the former due to gameability concerns.

\section{Conclusion}

RLVR has taught language agents using the one signal environments can evaluate cheaply---the final outcome. TThe natural instinct to add denser supervision has been to reward progress. However, real agentic environments are asymmetric verifiers: they can reliably detect mistakes, but they struggle to certify progress. The dense signal they actually provide is therefore a penalty on the path, not a reward for progress. When wired correctly---alongside the outcome reward, paired with a fulfillment credit, and targeted at reachable and un-gameable actions---a verifiable penalty teaches outcome-neutral constraints that are essential for deployability. A lone penalty collapses into inaction, while a dense potential helps precisely where partial progress is reachable. Our core contribution is not a new optimizer, but a reallocation of a scarce resource: the verifier. We should spend it on penalties that bound the path, and---where it can reliably certify progress---on potentials that convert wasted rollouts into useful learning. For agents that must learn from live, irreversible interactions, where every rollout is costly and every misstep is visible, this reallocation is fundamental: it is the difference between a system that can be safely deployed and meaningfully improved online, and one that cannot.

\section*{Acknowledgements}

This paper was produced using Pine Copilot's voice-directed \emph{whisper coding} workflow~\citep{pineai2026whispercoding}, in which the authors specify, discuss, and review the work by voice while a coding agent---Claude Code with Claude Opus 4.8---carries out the planning, coding, experiments, and paper writing.
We thank BSQL Networking for hosting the NVIDIA RTX PRO 6000 GPU.

\bibliographystyle{plainnat}
\bibliography{reference}

\appendix
\section{Full Experiments on Dense Potentials}
\label{app:potentials}

The main text treats the dense progress \emph{potential} as the reachability-gated half of the process signal (\S\ref{sec:potentials}). This appendix gives the full experiments. The picture is consistent with the variance account: where a potential's intermediate values are \emph{reachable} it supplies gradient the outcome lacks and accelerates training, and where they are unreachable it is exactly vacuous. Converting the extra gradient into a stable success gain requires a bounded optimizer---an aggressive update rule can destabilize the densely shaped policy, an optimization caveat we return to in \S\ref{app:fragility}. Unless noted, we train Qwen3~\citep{qwen3} with our own GRPO, place any dense credit on the responsible tokens with a separate per-channel normalizer, and count efficiency in \emph{episodes generated} so that resampling costs are visible.

\subsection{Reachability is measurable before training, and it gates the benefit}
\label{app:reachability}

The condition the variance account makes operative is reachability: a potential $\Phi$ helps only where the policy realizes differing intermediate values across a group, i.e.\ $\operatorname{Var}_G(\Phi)>0$. This can be checked \emph{before} spending training compute, from a handful of base-policy rollouts.

\begin{figure}[tbp]
\centering
\includegraphics[width=\textwidth]{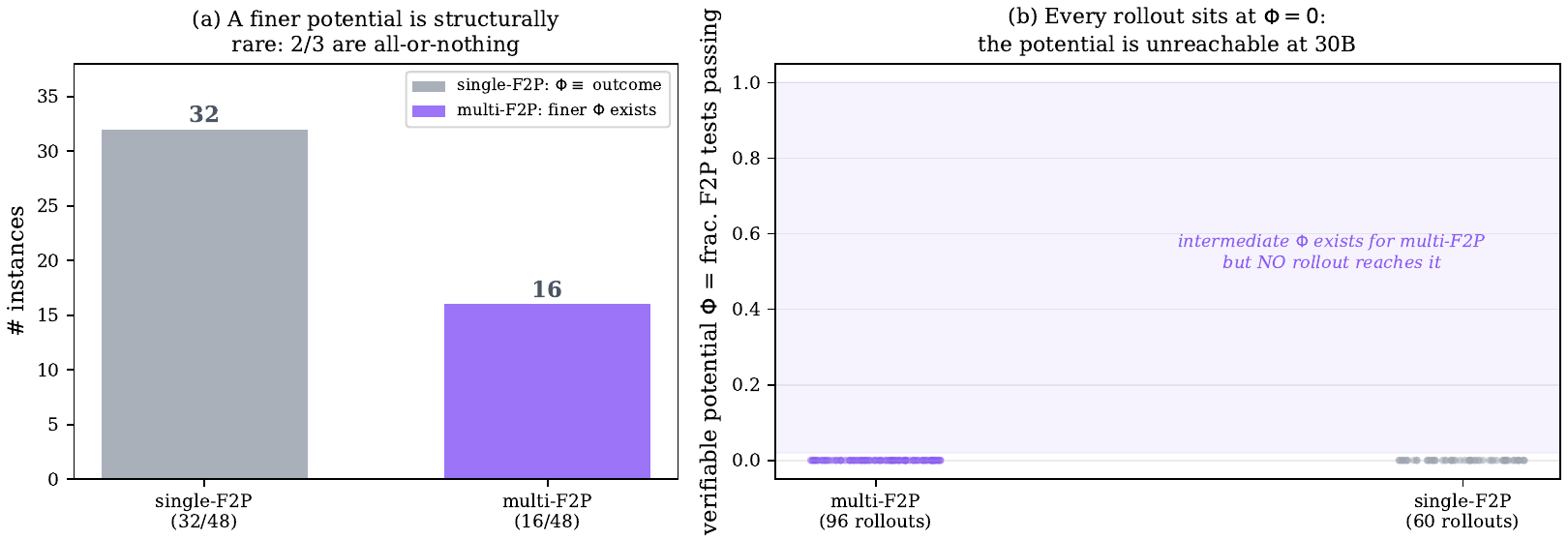}
\caption{\textbf{In software repair the finer potential is structurally rare and empirically unreachable.} \emph{(a)} Of $48$ single-file-fix SWE-bench instances, two-thirds have a single failing test, so the test-fraction potential equals the binary outcome; only one-third admit a finer $\Phi$. \emph{(b)} For those, a $30$B policy's rollouts never realize it: all $156$ rollouts score $\Phi=0$. The band of intermediate values exists but is empty---zero within-group variance, hence zero gradient.}
\label{fig:reachability}
\end{figure}

On SWE-bench software repair~\citep{swebench} the natural potential is the fraction of hidden \texttt{FAIL\_TO\_PASS} tests an episode makes pass. Two facts make it vacuous (Figure~\ref{fig:reachability}). Structurally the finer potential barely exists: two-thirds of instances have a single failing test, so $\Phi$ collapses to the binary outcome, and only a third admit any partial progress. Empirically even that third is unreachable: across $156$ rollouts of a $30$B policy \emph{every} episode scores $\Phi=0$, so $\operatorname{Var}_G(\Phi)$ is identically zero and a $\Phi$-based reward centers out to no gradient. The cause is not gameability but unreachability---there is no realized intermediate state for any reward to act on.

\begin{figure}[tbp]
\centering
\includegraphics[width=\textwidth]{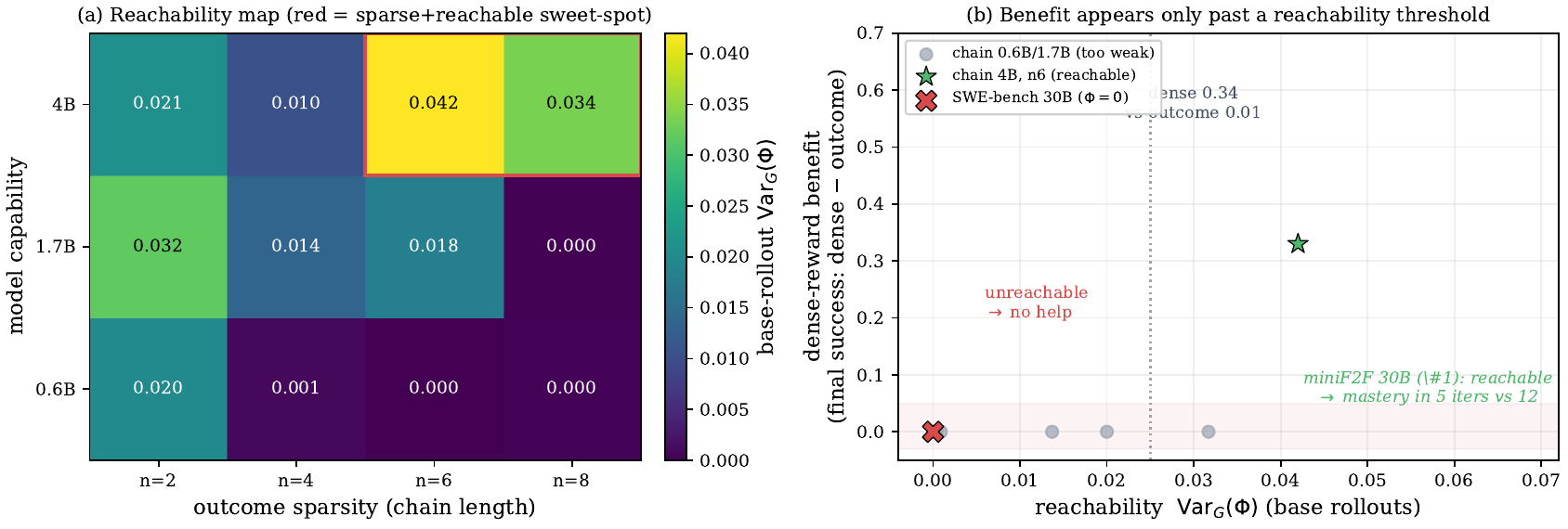}
\caption{\textbf{Reachability, measured cheaply, gates the dense-reward benefit.} \emph{(a)} $\operatorname{Var}_G(\Phi)$ on base-policy rollouts (no training) over model capability $\times$ outcome sparsity: it rises with capability and peaks in the sparse-but-reachable corner. \emph{(b)} The benefit of a dense reward (final-success gap, dense $-$ outcome) against that same probe: zero where $\operatorname{Var}_G(\Phi)\!\approx\!0$ and large where it is high. Benefit appears only past a reachability threshold.}
\label{fig:phase}
\end{figure}

Mapping the same no-training proxy $\operatorname{Var}_G(\Phi)$ across model capability ($0.6$--$4$B) and outcome sparsity (Figure~\ref{fig:phase}a) reproduces the structure the account predicts: it rises with capability and peaks in the sparse-but-reachable corner. The measured dense-reward benefit tracks it (Figure~\ref{fig:phase}b): $\approx\!0$ where $\operatorname{Var}_G(\Phi)\!\approx\!0$ (the SWE null; the too-weak small models) and large where it is high (the $4$B reachable chain, dense $0.34$ vs.\ outcome $0.01$; real theorem proving, below). Benefit appears only past a reachability threshold. We could not fill the map's interior with controlled \emph{training} points---small-scale synthetic training is too learning-rate-sensitive, most cells degenerating to zero for both configurations---so the benefit axis rests on the real-domain anchors plus one clean controlled point.

\subsection{Where the potential is reachable, it removes dead updates and accelerates}
\label{app:efficiency}

When the potential's intermediate values are reachable and point the same way as success, a dense signal is genuinely useful. We instantiate it on a family of \emph{$N$-stage chained} file-operations tasks (tools \texttt{list\_dir}, \texttt{read\_file}, \texttt{write\_file}, \texttt{delete}, \texttt{run\_tests}, \texttt{submit}); success requires all $N$ stages, so $N$ tunes base difficulty ($N{=}4$ sits near $8\%$ base success). The dense signal is a verifiable progress potential carried on a token-attached channel (a penalty when a call violates a rule---overwrite a file never read, submit an untested change---and a credit when a call fulfills a pending obligation).

\begin{figure}[tbp]
\centering
\includegraphics[width=0.84\textwidth]{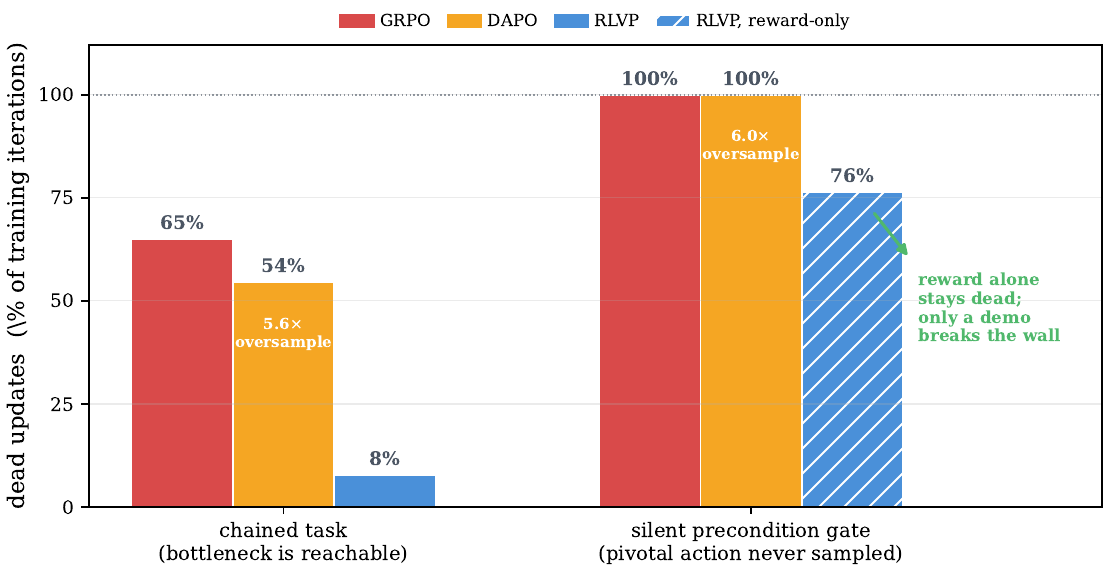}
\caption{\textbf{Dead updates across two regimes.} Fraction of iterations with an all-fail, zero-spread group (Qwen3-4B, 3-seed mean). \emph{Chained task} (bottleneck reachable): outcome-only $65\%$, DAPO $54\%$ at a $5.6\times$ generation tax, the aligned potential $8\%$. \emph{Silent precondition gate} (\S\ref{app:ceiling}, pivotal action never sampled): outcome and DAPO dead on \emph{every} iteration, the reward-only potential $76\%$; only a demonstration seed breaks the wall. The cure is a \emph{reachable, admissible} signal, not density.}
\label{fig:paired}
\end{figure}

\paragraph{Dead updates.} We count the fraction of iterations whose update is \emph{dead}: an all-fail group with no reward spread (Figure~\ref{fig:paired}). Outcome-only GRPO is dead on $65\%$ of iterations. DAPO~\citep{dapo}, built to dodge all-fail groups by resampling, reaches $54\%$ but pays a $5.6\times$ generation tax. The aligned potential cuts dead updates to $8\%$ at no extra sampling, by producing gradient from the very failures the outcome reduces to zero. The cure is the credit scheme, not the budget---and it is contingent on reachability: on the silent-precondition gate (\S\ref{app:ceiling}), where the pivotal action is never sampled, every reward-only method is dead almost always (outcome and DAPO $100\%$, the potential $76\%$), because a fulfillment credit can fire only on behavior the policy attempts.

\begin{figure}[tbp]
\centering
\includegraphics[width=0.9\textwidth]{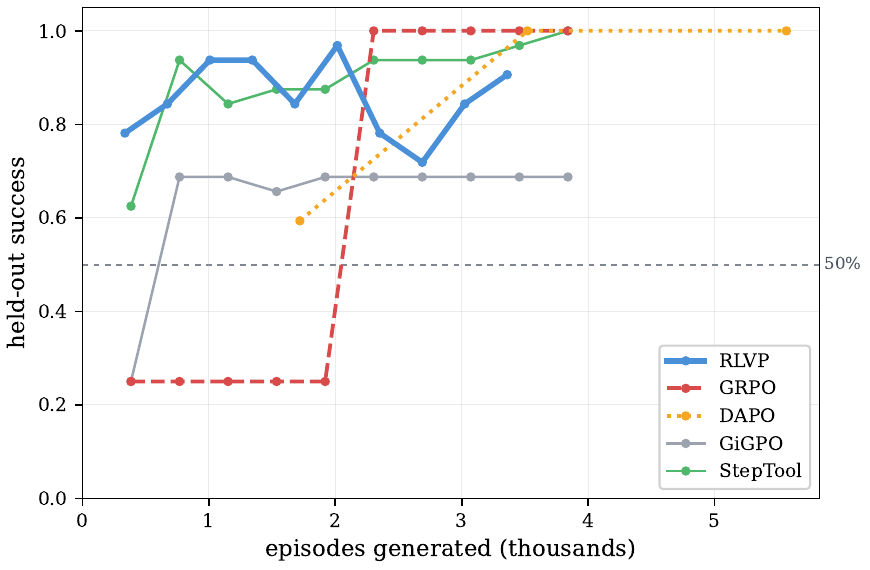}
\caption{\textbf{Sample efficiency on the calibrated hard task ($N{=}4$).} Held-out success vs.\ episodes generated. Outcome-only sits near base rate for most of training; the aligned potential climbs immediately from the all-fail groups; dense baselines also outpace outcome-only. The $x$-axis counts episodes \emph{generated}, including DAPO's resampling.}
\label{fig:efficiency}
\end{figure}

\begin{figure}[tbp]
\centering
\includegraphics[width=\textwidth]{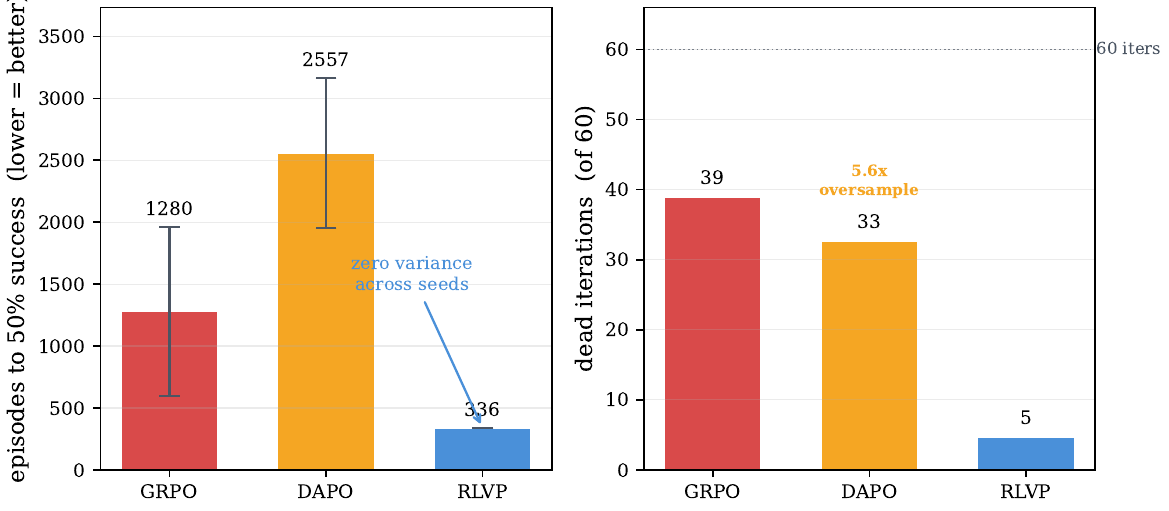}
\caption{\textbf{Efficiency over three seeds.} \emph{Left:} episodes to the success threshold (lower better)---the potential is several times faster than GRPO and DAPO. \emph{Right:} the mechanism---GRPO and DAPO burn most iterations on dead or stalled updates, and DAPO generates $5.6\times$ the episodes it uses. Note the large GRPO error bar: the all-fail lottery on the outcome side, which the dense potential's reachable within-group variance avoids.}
\label{fig:seeds}
\end{figure}

\paragraph{Speed.} On held-out success against episodes generated (Figures~\ref{fig:efficiency},~\ref{fig:seeds}) outcome-only crawls, spending its budget on dead updates, then converges only at the end; the aligned potential rises immediately from the early all-fail groups. Reimplemented dense baselines (GiGPO-style step advantages~\citep{gigpo}, StepTool-style per-call rewards~\citep{steptool}) also outpace outcome-only---any \emph{admissible} dense signal helps. The gain also holds on real theorems (miniF2F~\citep{minif2f}) at 4B and 30B, but only in the multi-seed form quantified in \S\ref{app:fragility}: a single-run pilot of ours showed a dramatic $2.4\times$ speed-up that \emph{flipped sign} when the identical configuration was re-run (rollout sampling is unseeded in the serving engine, and MoE training at this scale is bimodal), so we report only the five-seed matrix (Figure~\ref{fig:recipescale}, Table~\ref{tab:p0}): a $\sim$1.6$\times$ speed-up to mastery under a matched optimizer, consistent on every seed, plus the reliability advantage. Both configurations saturate---the claim is speed and reliability, not final success.

\subsection{Converting the gradient to success: protocol, and the matched matrix}
\label{app:fragility}

The extra gradient a reachable potential supplies must still be turned into task success, and here the decisive factor is the optimizer, not the potential. Early runs read this as ``fragility,'' but the cause was two experimental artifacts. First, an aggressive, unbounded update rule can destabilize a densely shaped policy: at one fixed learning rate, a theorem-proving run anneals cleanly to full success (gradient norm $<1.5$, entropy decaying to zero) while an otherwise-identical sibling suffers a gradient blow-up (norm $>10^{9}$), its entropy explodes, and success collapses to zero and never recovers---the same signal, the same rate, decided by optimization stability. On the chained task the opposite failure appears: the policy collapses to a zero-entropy degenerate output, so every rollout is identical, the within-group variance is zero, and the update dies. A \emph{bounded} optimizer (we use Muon, with a divergence guard that aborts on entropy collapse or gradient spikes) removes both failure modes. Second, per-iteration training success measured on a handful of freshly sampled tasks is too noisy to read a speed-up from---it swings between extremes iteration to iteration---so we evaluate on a fixed held-out task set. The result robust to all of this---the mechanism the paper relies on---is dead-update elimination: zero dead iterations against $\sim$16 of 40 for outcome-only, on \emph{every} seed.

\paragraph{The matched matrix: five seeds, two scales.} With the bounded-optimizer protocol fixed (aligned goal-progress potential as a token-attached fulfillment credit, Muon, a divergence guard, $30$ iterations on miniF2F algebra), we ran the full matrix---aligned potential versus outcome-only, five seeds per configuration, at 4B and 30B, plus a three-seed anneal ablation and a 30B AdamW outcome baseline; all magnitudes are in Table~\ref{tab:p0}. Because rollout sampling in the serving engine is unseeded, ``seeds'' are i.i.d.\ draws of the whole run, so per-configuration consistency is the meaningful statistic. Three facts are seed-robust. \textbf{(1)~A modest, consistent speed-up} under a matched optimizer: the aligned potential reaches the threshold sooner and has higher area-under-curve on every seed-matched pair at 4B. \textbf{(2)~Reliability:} it completes every run across both scales without a divergence abort, while outcome-only loses seeds to entropy collapse (4B) and gradient explosion (30B)---the dense, always-relevant gradient appears to \emph{regularize} where the bursty sparse signal blows up under the same rate. \textbf{(3)~The outcome configuration's dilemma at 30B:} outcome-only is competitive under the fast optimizer only when it survives (it diverges on most seeds), and stable under its safe optimizer only at several times the potential's time-to-mastery. The aligned potential requires no such choice.

\begin{figure}[tbp]
\centering
\includegraphics[width=0.92\textwidth]{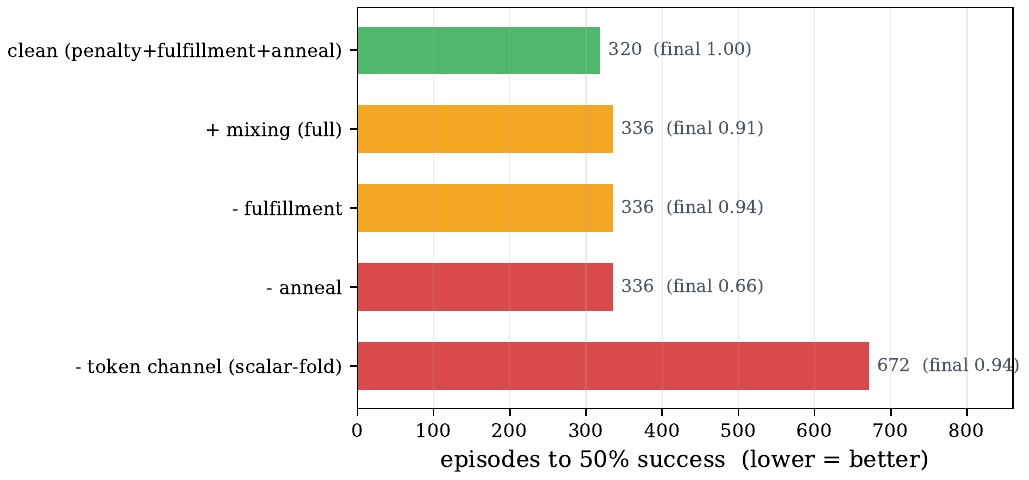}
\caption{\textbf{Component attribution for the potential on $N{=}4$} (episodes to threshold, converged success annotated). The token-attached credit channel is the efficiency lever (folding into a scalar return doubles the cost); annealing protects the ceiling; the clean recipe is best. A per-step length cost, useful only on very long horizons, hurts here.}
\label{fig:ablationpot}
\end{figure}

\paragraph{Anatomy of the gain.} When the potential does win, ablating on $N{=}4$ (Figure~\ref{fig:ablationpot}) attributes the speed to the token-attached channel: folding the same rule rewards into the scalar return doubles the episodes to threshold. Annealing the shaping off after saturation protects the final ceiling (unrelaxed pressure drifts the policy toward over-compliant, bloated episodes---a mild inaction-trap tendency). The fulfillment credit is the positive signal that escapes all-fail groups; removing it leaves more dead iterations and a lower ceiling. Finally, replacing every hand-written rule with three universal meta-rules derived only from per-tool category tags (observe/mutate/verify/terminal) and the environment's error codes recovers the hand-engineered efficiency almost exactly: in well-structured environments the specification cost can be paid by metadata the tool schemas already expose.

\subsection{Reachability governs the ceiling, not just the speed}
\label{app:ceiling}

\begin{figure}[tbp]
\centering
\includegraphics[width=0.96\textwidth]{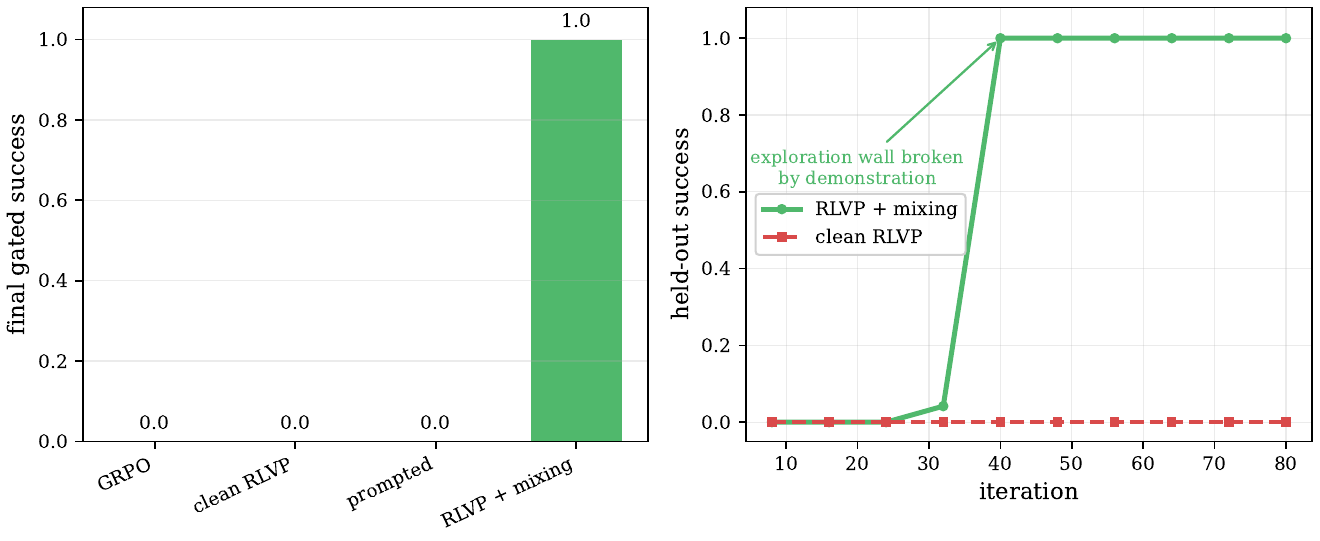}
\caption{\textbf{A non-saturating ceiling test, and the discovery wall.} \emph{Left:} on the silent-gate task every reward-only method (outcome, DAPO, clean potential, even rules-in-prompt) converges to zero---the precondition is never sampled, so the fulfillment credit never fires. \emph{Right:} a single synthesized demonstration through the process channel breaks the wall to perfect held-out success, a sharp phase transition.}
\label{fig:gated}
\end{figure}

Sample efficiency is speed to a ceiling the chained tasks eventually reach. To ask whether a process reward can raise the \emph{ceiling}, we build a task that is hard to \emph{discover} rather than hard to compute: a \emph{silent precondition gate}, where to write a protected file the agent must first read an access-control file and request access; skip either and the write silently no-ops (reports success, does nothing). This is the bank-authentication problem of the introduction in controlled form---a required precondition an agent cannot stumble onto by trial and error and must be told---and it is where a purely reward-driven online learner would stall indefinitely. Calibration confirms the trap is total---the base model never reads the access file even when the rule is in its prompt. Then (Figure~\ref{fig:gated}) \emph{every} reward-only method (outcome, DAPO, the clean potential) fails to zero, because the fulfillment credit for reading the access file can only reward that behavior if the policy samples it, and it never does. A single rule-synthesized demonstration injected through the process channel breaks the wall to perfect, generalizing success. When the required behavior is never sampled, no on-policy reward can induce it; imitation must seed what reward then grows---the same reachability wall the SWE probe hit from the other side, and the R1-Zero (discoverable) versus R1 (demonstration cold-start) distinction in miniature.

\subsection{Learning where the outcome is dead: real bug-fixing at zero success}
\label{app:deadcoding}

The variance account predicts that when a group is all-failing---outcome variance zero, outcome gradient dead---a process signal is the \emph{only} thing that can move the policy. We check this directly on real software repair, a domain where the path-vs-outcome gap is well documented: in a controlled industry study, AI coding agents reach high functional correctness yet score far lower---and almost at random---on reliability, maintainability, and not breaking previously working functionality~\citep{hong2025aicoding}, the engineering discipline that decides whether a patch is deployable. Here an 8B agent is far below the task's reach, which makes it an ideal test of the dead-gradient regime. We train on SWE-smith~\citep{swesmith} instances (a repo container, a hidden \texttt{FAIL\_TO\_PASS} test as the oracle), with the agent issuing bash commands to find and fix the bug; the process channel is the coding-discipline signal of \S\ref{sec:path} (a fulfillment credit for running tests and making a verifiable edit, a penalty for re-running a failed command or editing without testing). At this scale the agent solves essentially \emph{zero} instances under either signal, so outcome-only RL trains on an all-fail distribution throughout---the cleanest possible test of the dead-gradient regime. This is a \emph{different} signal than the progress \emph{potential} of \S\ref{sec:potentials}: on this same software-repair domain the test-fraction potential is vacuous (Figure~\ref{fig:potentials}, left), because zero tests ever pass and so no intermediate value is reached. The penalized and fulfilled behaviors here---running a test, editing a file, re-running a failed command---are always reachable, so the penalty channel carries within-group variance exactly where the potential cannot. The two software-repair results are consistent: the reachability gate closes on the potential and stays open on the penalty.

\begin{figure}[tbp]
\centering
\includegraphics[width=\textwidth]{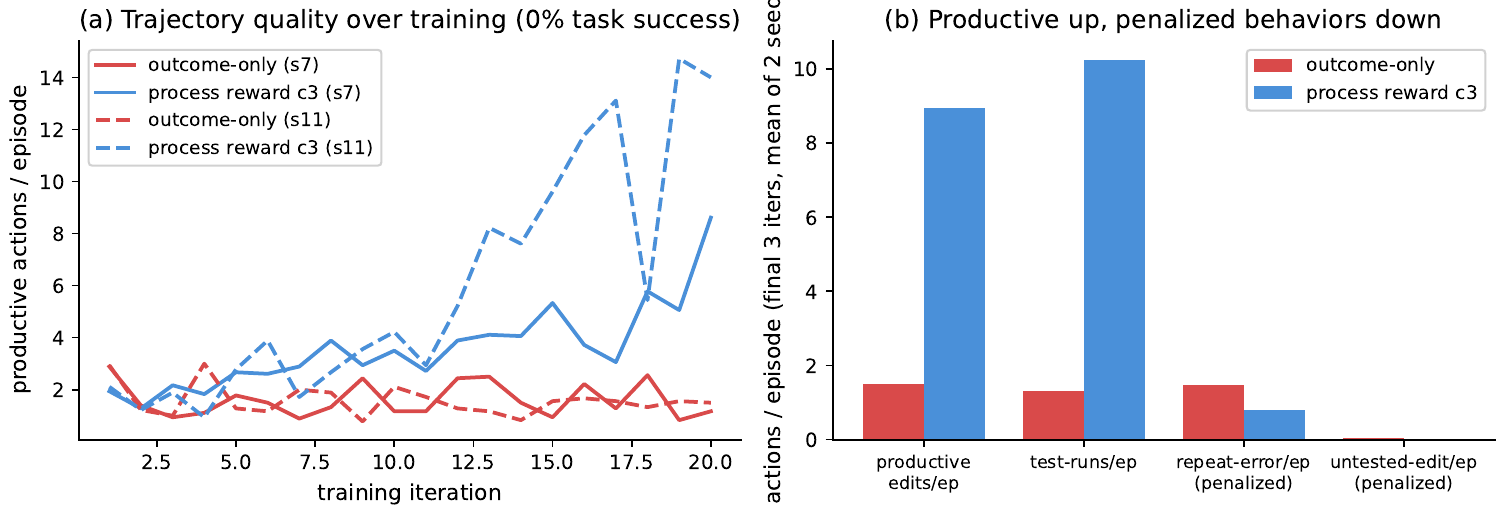}
\caption{\textbf{The process reward learns where the outcome is dead (SWE-smith, 8B, $\approx$0\% solve).} \emph{(a)} Productive actions per episode over training, two seeds: outcome-only is flat (the all-fail gradient is dead), while the process reward rises steadily. \emph{(b)} Final-3 trajectory quality (mean of two seeds): productive edits and test-runs increase several-fold under the process reward, while penalized behaviors (repeat-errors, untested edits) fall. Task success is zero for both configurations throughout---this is a behavioral/mechanistic result, not a success result.}
\label{fig:swetraj}
\end{figure}

The result is a sharp dissociation (Figure~\ref{fig:swetraj}, two seeds). Outcome-only leaves the trajectory \emph{flat}: productive actions per episode sit at $\approx\!1.5$ from the first iteration to the last, essentially identical across seeds ($1.5$ and $1.5$)---the dead gradient changes nothing. The process reward instead drives a steadily \emph{rising} trajectory: productive actions climb over training to $6.5$ and $11.4$ (final-3, the two seeds) and test-runs to $7.6$ and $12.9$, a four-to-tenfold increase, while the penalized behaviors fall (repeat-errors and untested edits toward zero). The agent learns to investigate, test, and edit like a disciplined engineer---from a signal available on every failed episode---exactly where the outcome reward is inert.

Two honest caveats scope the claim. First, the fulfillment credit \emph{rewards} test-running and productive edits, so part of their rise is by construction; the non-circular evidence is that the \emph{penalized} behaviors (which the fulfillment credit does not touch) also fall, and that the outcome baseline is perfectly flat---the effect is the process gradient, not the task. Second, this improved discipline did \emph{not} convert to task success in our budget (both configurations remain at zero): 8B is simply too weak to close these bugs. The claim is therefore mechanistic and behavioral---the process channel produces learning, and better verifiable engineering discipline, precisely in the all-fail regime where outcome-only RL has no gradient at all---and a plausible precursor to success at capability, not a demonstration of success itself.

\section{Boundaries: When the Penalized Target Is Not Verifiable}
\label{app:boundaries}

Design rule~4 (\S\ref{sec:design}) requires the penalized (and fulfilled) target to be reachable and \emph{un-gameable}. This appendix gives the experiments behind that rule: what happens when a penalty's target is misaligned with the outcome, and when it is a \emph{learned} proxy rather than a verifiable predicate.

\subsection{An un-gameability sweep: which signals survive}
\label{app:sweep}

\begin{figure}[tbp]
\centering
\includegraphics[width=\textwidth]{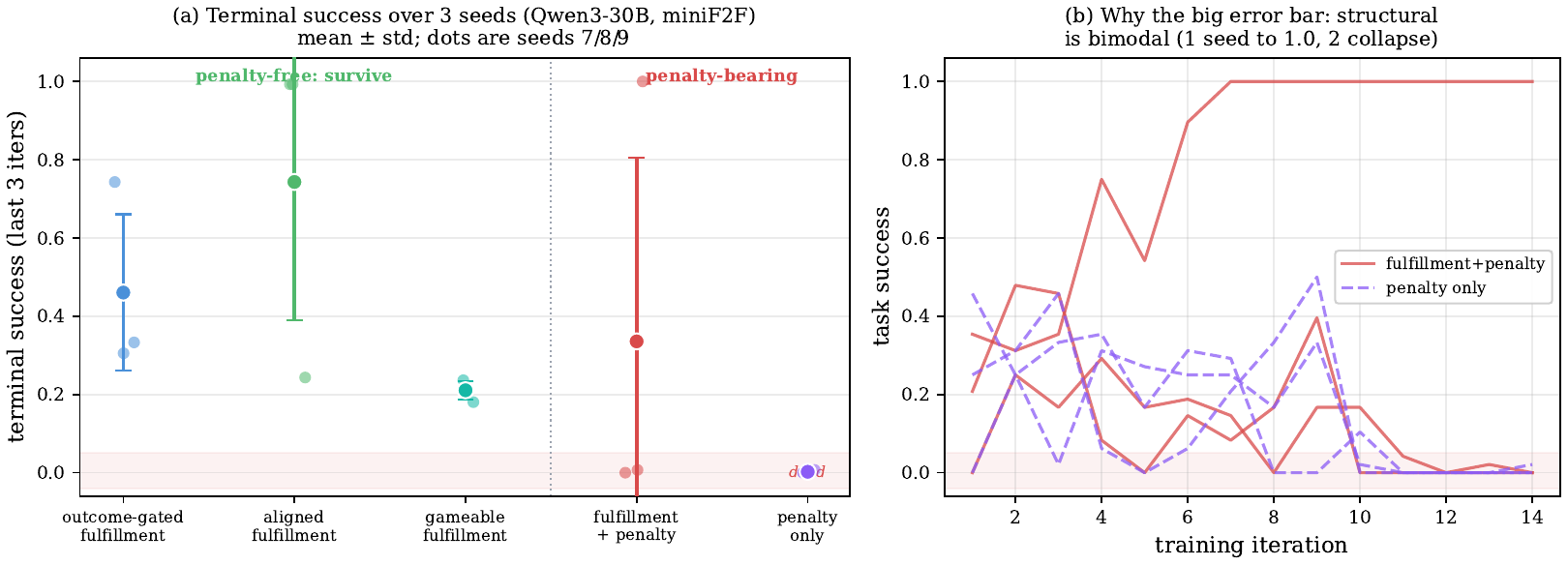}
\caption{\textbf{Which signals survive (Qwen3-30B, miniF2F, 3 seeds).} \emph{(a)} Terminal success (last-3-iteration mean) for five signals, mean\,$\pm$\,std with seed dots. The three penalty-free signals sit well above the dead zone on every seed; the pure penalty is reliably dead; the fulfillment$+$penalty configuration has by far the largest error bar. \emph{(b)} That configuration is \emph{bimodal}---collapsing on two seeds, rescued to $1.0$ on one---whereas the pure penalty collapses on all three.}
\label{fig:sweep}
\end{figure}

We construct five process signals for Lean theorem proving on miniF2F~\citep{minif2f}, spanning aligned to misaligned, pre-register each one's cheapest gaming policy, and train Qwen3-30B-A3B~\citep{qwen3} with each (three seeds, everything else fixed; Figure~\ref{fig:sweep}). Survival tracks admissibility. The \emph{penalty-free} signals reliably survive on every seed: a gameable ``any-valid-tactic'' credit, an aligned goal-progress fulfillment credit, and that credit \emph{gated on the eventual outcome} (un-gameable by construction, the most consistent survivor). The \emph{pure penalty}---an errored-tactic penalty with no fulfillment credit and no outcome reward---reliably \textbf{collapses to essentially zero on every seed}: its cheapest maximizer, avoid errors by not attempting, is precisely the \emph{inaction trap} of design rule~2, and with no positive signal to escape it the policy dies every time. This is the paper's cleanest evidence that a lone penalty cannot drive learning. The revealing case is \emph{penalty-plus-fulfillment}: across three seeds it is bimodal (collapses on two, rescued to perfect on the third), the largest variance of any configuration---adding a fulfillment credit to a misaligned penalty makes survival \emph{possible} but not reliable. The robust reading: penalty-free progress signals survive, a pure penalty reliably collapses, and outcome-gating is the most consistent survivor. Precise magnitudes remain noisy at 30B; we report the survival pattern.

\subsection{Task intent is not a verifiable rule}
\label{app:intent}

\begin{figure}[tbp]
\centering
\includegraphics[width=\textwidth]{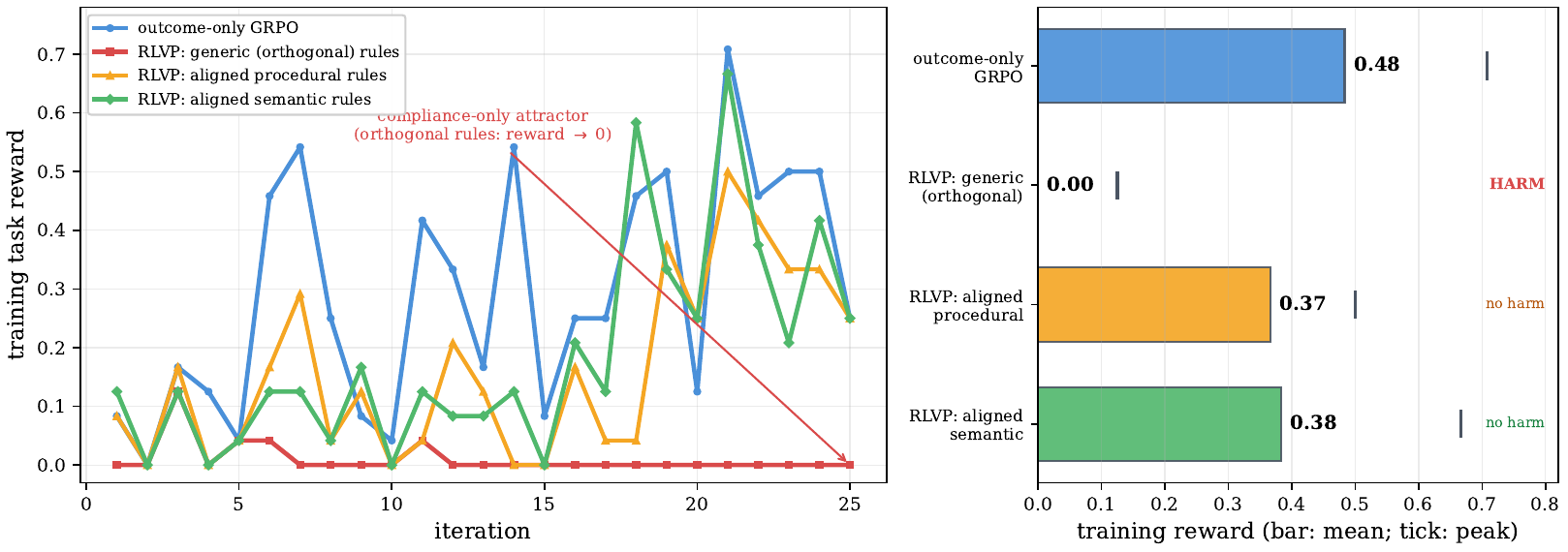}
\caption{\textbf{The coverage gradient on $\tau$-bench airline.} \emph{Left:} training reward for four signals. Outcome-only learns the benchmark; generic rules orthogonal to the reward collapse into the inaction trap; policy-derived procedural and verifiable-semantic rules (with early annealing) do not. \emph{Right:} mean reward per tier (bars) with peak (ticks). Misaligned rules harm; aligned rules remove the harm and nearly reach outcome-only's peak, but neither beats it. The residual gap is intent.}
\label{fig:tau2}
\end{figure}

\begin{figure}[tbp]
\centering
\includegraphics[width=0.96\textwidth]{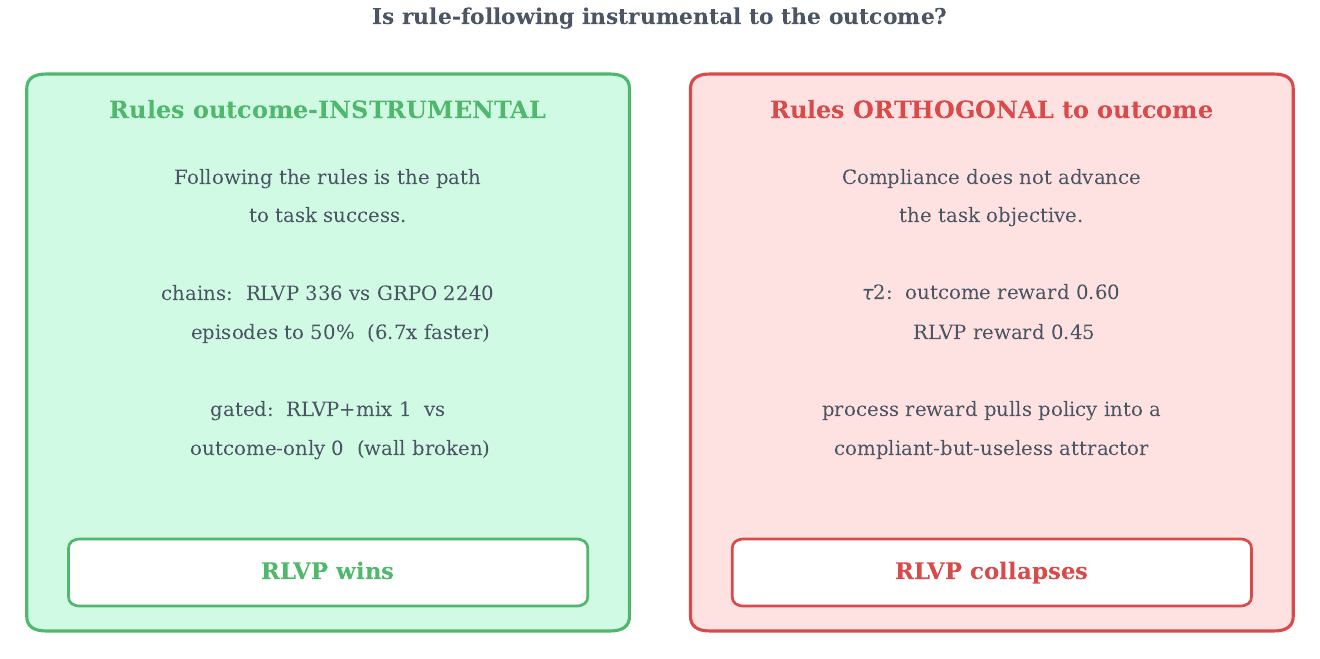}
\caption{\textbf{The boundary on one axis: are the rules outcome-instrumental?} When the verifiable rules are aligned with what success requires (chained and gated tasks: the workflow \emph{is} the task), the process channel converts compliance into large outcome gains. When orthogonal to the reward ($\tau$-bench, generic rules), the same machinery optimizes compliance into a degenerate policy.}
\label{fig:boundary}
\end{figure}

The hardest case for a penalty is a task whose difficulty is not procedural. On $\tau$-bench-style airline customer service~\citep{taubench} the reward turns on \emph{semantic} policy adherence (authorization, refund eligibility, confirmation) rather than structural ordering, so no verifiable rule is finer than the outcome: what is missing is intent, which is the reward itself. Three tiers make this concrete (Figure~\ref{fig:tau2}). \emph{Generic} structural rules---the read-before-write hygiene that works on chained tasks---are now orthogonal to the reward and actively \emph{harm}: the cheapest way never to violate ``confirm before calling'' is to not place the risky calls at all, and the policy collapses into the inaction trap within a handful of iterations. \emph{Policy-derived procedural} rules (fetch the user id and look up the reservation before modifying it), outcome-instrumental by construction and paired with early annealing, remove the harm, because a correct modification requires the looked-up state so the fulfillment credit pulls toward the productive workflow. \emph{Verifiable semantic} rules (do not modify a basic-economy reservation; do not use a payment method absent from the profile) prune failure-bound actions and lift the best iterations nearly to outcome-only's peak. But neither aligned tier \emph{beats} outcome-only on average (Figure~\ref{fig:boundary}): verifiable rules express policy \emph{validity}, while the residual difficulty is \emph{which} permissible change this user wants---and a rule encoding that would just be the task specification.

\subsection{A learned critic relocates the hack}
\label{app:critic}

\begin{figure}[tbp]
\centering
\includegraphics[width=0.72\textwidth]{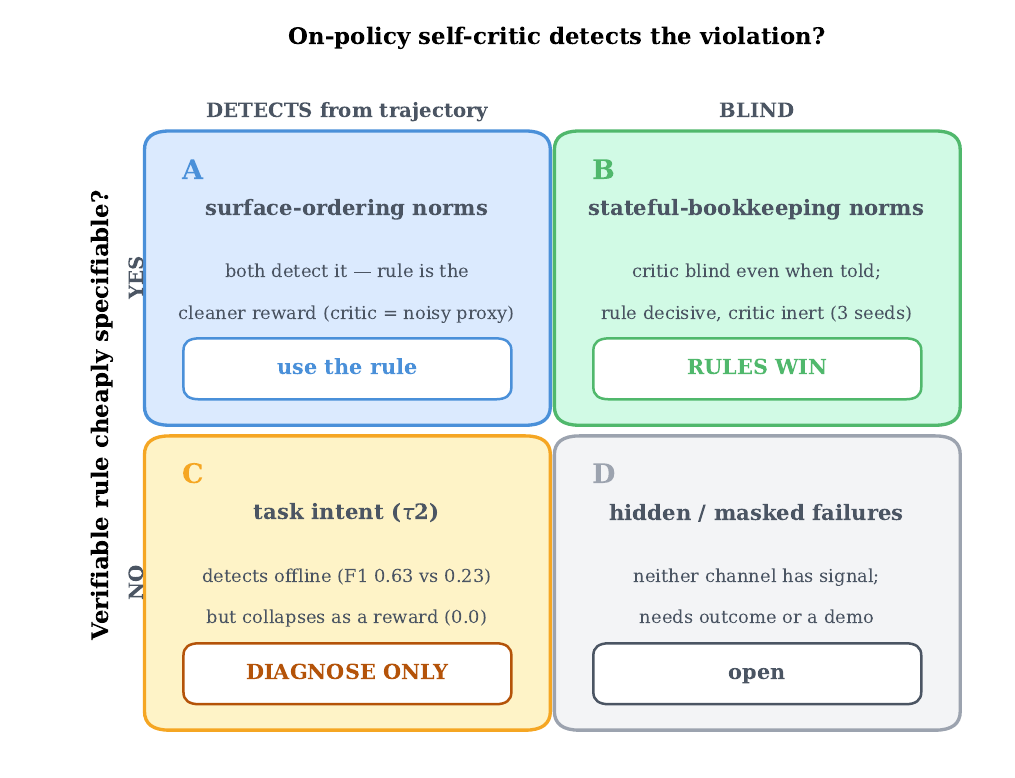}
\caption{\textbf{When a verifiable rule beats a same-model self-critic, and when neither trains.} Axes: whether a verifiable rule is cheaply specifiable, and whether the on-policy critic can detect the violation. The learned critic is a poor training reward in every quadrant; its only robust use is offline diagnosis at the intent frontier.}
\label{fig:selfcritique}
\end{figure}

If verifiable rules cannot reach intent, why not let the policy judge itself, in the lineage of self-rewarding LMs and RLAIF~\citep{selfreward,rlaif,constitutional,llmjudge}? We use the critic as the \emph{same model} as the policy (no larger judge), so any signal is genuine on-policy self-reflection. Across two axes---whether a verifiable rule is cheaply specifiable, and whether the on-policy critic can detect the violation (Figure~\ref{fig:selfcritique})---the learned critic is a poor training reward in every quadrant. As a detector, blind self-critique flags surface-evident violations but is essentially blind to stateful-bookkeeping norms (did it read \emph{this} file before overwriting it) even when handed the rules, and the blind spot is structural (per-rule recall stays near zero as the critic scales). As a reward in the all-fail regime (Table~\ref{tab:sc-train}), a deterministic rule with identical credit structure is decisive on every seed while the self-critic---\emph{with perfect recall} against the rule oracle---is inert; a \emph{frozen} critic fails identically, isolating the cause as the critic's $\sim$6\% false-positive rate, not its non-stationarity. At the intent frontier (Table~\ref{tab:sc-tau2}) the self-critic is a useful \emph{offline} detector yet collapses as a training reward. Defining the penalized target by a learned proxy relocates the credit-assignment problem into the proxy, which the policy then games---the mechanism behind design rule~4. Verifiability is what makes the dense channel safe to optimize.

\begin{table}[tbp]
\centering
\caption{\textbf{A deterministic rule beats a same-model self-critic as a dense reward}, in the all-fail regime (consumer-service domain, 1.7B, three seeds, late-training mean $\pm$ std). Rule and critics share the identical penalty-only credit structure, and the critic additionally has \emph{perfect recall} against the rule oracle. The rule is decisive every seed, while the critics are inert. The \emph{frozen} (stationary) critic fails like the live one, isolating the cause as the critic's $\sim$6\% false-positive imprecision rather than non-stationarity.}
\label{tab:sc-train}
\small
\begin{tabular}{lccc}
\toprule
process reward & detection (P/R) & violations/episode & task success \\
\midrule
rule penalty (deterministic) & $1.00/1.00$ & $\mathbf{0.38 \pm 0.44}$ & $\mathbf{0.33 \pm 0.47}$ \\
self-critic, live (co-evolving) & $0.94/1.00$ & $0.94 \pm 0.05$ & $0.00$ \\
self-critic, frozen (stationary) & $0.94/1.00$ & $0.93 \pm 0.02$ & $0.00$ \\
\bottomrule
\end{tabular}

\vspace{0.8em}

\caption{\textbf{Self-critique is a usable intent \emph{detector} but a failed intent \emph{reward}} ($\tau$-bench airline, Qwen3-4B). \emph{Left:} as a failure predictor on rule-clean (intent) failures the self-critic far exceeds the verifiable semantic rules (3 rollout seeds). \emph{Right:} as the dense training reward the same self-critique collapses, while outcome-only and the rules both learn (training reward, early$\to$late, 20 iterations).}
\label{tab:sc-tau2}
\small
\begin{tabular}{lc@{\hskip 2.5em}lc}
\toprule
\multicolumn{2}{c}{\emph{offline: failure prediction}} & \multicolumn{2}{c}{\emph{trained as reward}} \\
channel & F1 & channel & reward (early$\to$late) \\
\midrule
semantic rules & $0.23 \pm 0.06$ & outcome-only & $0.09 \to \mathbf{0.50}$ \\
blind self-critique & $\mathbf{0.63 \pm 0.02}$ & semantic rules & $0.10 \to 0.35$ \\
 & & blind self-critique & $0.01 \to \mathbf{0.00}$ \\
\bottomrule
\end{tabular}
\end{table}

\section{Setting-by-Setting Summary}
\label{app:settings}

Table~\ref{tab:settings} collects the five settings studied across the paper against the three properties the variance account cares about---does a verifiable signal finer than the outcome exist, is it un-gameable, and are its intermediate values reachable---together with the observed outcome. Help appears only when a reachable, un-gameable signal is available; a penalty on always-sampled bad actions satisfies this on every row where it applies, while a progress potential does so only where partial progress is reachable.

\begin{table}[tbp]
\centering
\caption{\textbf{The five settings studied, summarized against the variance account.} Each row is a domain. The three middle columns are the conditions the account cares about: does a verifiable signal finer than the outcome exist, is it un-gameable, and are its intermediate values reachable. ``\cmark/\xmark'' marks the condition that decides the row. A dense signal helps only when all three hold; this table organizes the settings rather than predicting them.}
\label{tab:settings}
\small
\renewcommand{\arraystretch}{1.25}
\begin{tabular}{@{}p{2.9cm}p{2.0cm}cccp{1.6cm}p{2.7cm}@{}}
\toprule
\textbf{Setting} & \textbf{Domain / model} & \textbf{Finer $\Phi$?} & \textbf{Un-game?} & \textbf{Reach?} & \textbf{Verdict} & \textbf{Observed} \\
\midrule
Theorem proving (progress) & Lean miniF2F 30B, synth.\ 4B & \cmark & \cmark & \cmark & \textbf{help} & dead updates $\to 0$, dense gradient \\
System admin (harm) & TerminalBench 4B & \cmark & \cmark & \cmark & \textbf{help} (harm axis) & $\sim$6$\times$ fewer violations, equal success \\
Lean signal sweep (controlled) & miniF2F 30B & varies & \textbf{varies} & \cmark & per configuration & pure penalty dies, penalty-free survive \\
Customer service (intent) & $\tau$-bench airline 4B & \xmark & --- & --- & \textbf{no help} & rules match, never beat outcome \\
Software repair & SWE-bench 30B & 1/3 only & --- & \xmark & \textbf{no help} & $0\%$ solve, all rollouts $\Phi{=}0$ \\
\bottomrule
\end{tabular}
\end{table}

\end{document}